\documentclass[10pt,journal]{IEEEtran}
\IEEEoverridecommandlockouts 
\usepackage{blkarray}                                      
\usepackage{algpseudocode}                                 
\usepackage{algorithm}
\algblockdefx[pardo]{pardo}{endpardo}%
   [1]{\textbf{For} #1 }%
   {\textbf{Endfor}}
\usepackage{graphicx}                                      
\usepackage{amsmath}
\usepackage{amssymb}
\usepackage{amsfonts}
\usepackage{amsthm}
\usepackage[mathcal]{eucal}
\usepackage{mathrsfs}
\usepackage{booktabs}
\usepackage{enumerate}
\usepackage{multirow}
\usepackage[subrefformat=parens,farskip=0pt,justification=centering]{subfig}
\usepackage{color}
\usepackage{cite}                                          
\usepackage{comment}                                       
\usepackage{soul}                                          
\soulregister\cite7
\soulregister\ref7
\soulregister\pageref7
\usepackage{etoolbox}                                      
\usepackage{url}
\usepackage{nth}                                           
\usepackage{bm}                                            
\usepackage{courier}
\usepackage{balance}
\usepackage{threeparttable}
\usepackage{xcolor,colortbl}
\usepackage{footnote}
\usepackage{listings}
\usepackage{setspace}                                      

\usepackage[frozencache,cachedir=.]{minted}

\usepackage{verbatim}
\usepackage[bookmarks=false]{hyperref}
\hypersetup{
    colorlinks = true,
    citecolor  = blue,
    linkcolor  = blue,
    urlcolor   = blue,
}
\usepackage{tikz}
\usetikzlibrary{patterns,snakes}
\usetikzlibrary{positioning,calc,fit,decorations.pathmorphing,shapes.geometric, shapes.gates.logic.US, calc}
\usetikzlibrary{arrows,arrows.meta,decorations.markings,shapes,shapes.arrows}
\usetikzlibrary{decorations,decorations.pathreplacing}
\usetikzlibrary{backgrounds}
\usepackage{filecontents}                                  
\usepackage{pgfplots}
\usepackage{pgfplotstable}
\usepackage{scalefnt}
\pgfplotsset{compat=newest}
\usepackage{caption}
\usepackage{pifont}                                        
\usepackage{cleveref}
\Crefformat{figure}{Fig.~#2#1#3}                           
\Crefname{subfigure}{Fig.}{Figs.}
\Crefname{figure}{Fig.}{Figs.}
\Crefformat{table}{TABLE~#2#1#3}                           
\usepackage[figuresright]{rotating}

\definecolor{CUHKorange}{RGB}{244,106,18} 
\definecolor{CUHKblue}{RGB}{0,111,190}    
\definecolor{CUHKgreen}{RGB}{0,127,128}   
\definecolor{CUHKred}{RGB}{228,46,36}     
\definecolor{CUHKyellow}{RGB}{198,148,34} 
\definecolor{CUHKdark}{RGB}{114,44,114}   
\definecolor{CUHKmiddle}{RGB}{144,44,144} 
\definecolor{CUHKlight}{RGB}{167,44,167} 
\definecolor{CUHKpurple}{RGB}{117,15,109}
\definecolor{CUHKgold}{RGB}{221,163,0}
\definecolor{CUHKribbon}{RGB}{244,223,176}
\definecolor{CUHKblack}{RGB}{34,24,21}

\definecolor{myboxbg}{RGB}{255,249,232}  
\definecolor{myboxframe}{RGB}{209,154,0} 

\usepackage[most]{tcolorbox}

\newcommand{\mycenterheading}[1]{%
  \par\vspace{0.8ex}
  {\centerline{\normalsize\textbf{#1}}}
  \vspace{0.8ex}\par
}
\usepackage{relsize}

\newcommand{\mysubheading}[1]{%
  \par\vspace{0.6ex}
  {\noindent\small\textbf{#1}}
  \vspace{0.4ex}\par
}

\renewcommand{\bf}[1]{\textbf{#1}}
\renewcommand{\tt}[1]{\texttt{#1}}

\usepackage{tcolorbox}

\tcbuselibrary{skins,breakable}
    {\endtcolorbox}
    {\endtcolorbox}

\usepackage[top=0.78in,bottom=0.86in,left=0.62in,right=0.62in]{geometry}
\iftrue
\newcommand{\subparagraph}{}
\usepackage{titlesec}
\titlespacing*{\section}{0pt}{1.8ex plus .2ex minus .2ex}{0.4ex plus .2ex}
\titlespacing*{\subsection}{0pt}{1.0ex plus .2ex minus .2ex}{0.2ex plus .2ex}
\fi

\crefname{mytheorem}{Theorem}{Theorems}
\crefname{mylemma}{Lemma}{Lemmas}
\crefname{myclaim}{Claim}{Claims}
\crefname{myproperty}{Property}{Properties}
\crefname{mycorollary}{Corollary}{Corollaries}

\algrenewcommand\textproc{\texttt}

\makeatletter
\let\OldStatex\Statex
\renewcommand{\Statex}[1][3]{%
  \setlength\@tempdima{\algorithmicindent}%
  \OldStatex\hskip\dimexpr#1\@tempdima\relax
}
\makeatother

\RequirePackage[normalem]{ulem} 
\RequirePackage{color}\definecolor{RED}{rgb}{1,0,0}\definecolor{BLUE}{rgb}{0,0,1} 

\definecolor{myblue}{rgb}{225, 235, 246}
\definecolor{myred}{rgb}{202, 223, 184}
\graphicspath{{./figs/}{../}}

\allowdisplaybreaks

\usepackage{adjustbox}
\usepackage{xcolor}
\definecolor{titleamber}{RGB}{215, 155, 0}
\definecolor{boxpaleamber}{RGB}{255, 250, 230}
\definecolor{orangetask}{RGB}{255, 140, 0}
\definecolor{emphcolor}{HTML}{1A5276}
\definecolor{icbest}{rgb}{0.85,0.9,1.0}
\definecolor{boxemph}{RGB}{20, 40, 120}
\definecolor{titleblueviolet}{RGB}{88, 80, 202}
\definecolor{boxpaleblueviolet}{RGB}{235, 235, 255}
\definecolor{easy_green}{HTML}{93C47D}
\definecolor{medium_orange}{HTML}{F6B26B}
\definecolor{hard_red}{HTML}{E06666}
\definecolor{lightGreen}{HTML}{D9EAD3}
\definecolor{lightOrange}{HTML}{FCE5CD}
\definecolor{lightRed}{HTML}{F4CCCC}

\newtcolorbox{promptbox}[1][]{
  enhanced,
  attach boxed title to top left={yshift=-2mm, xshift=0mm},
  colback=boxpaleamber,
  colframe=titleamber!70!white,
  boxed title style={colback=titleamber, colframe=titleamber, size=small, left=3mm, right=3mm},
  title=#1,
  fonttitle=\color{white}\bfseries,
  arc=2mm,
  boxrule=0.5pt,
  left=5pt,
  right=5pt,
  top=10pt,
  bottom=5pt,
  fontupper=\small,
  breakable
}

\newtcolorbox{promptboxmllm}[1][]{
  enhanced,
  attach boxed title to top left={yshift=-2mm, xshift=0mm},
  colback=boxpaleblueviolet,
  colframe=titleblueviolet!70!white,
  boxed title style={
    colback=titleblueviolet, 
    colframe=titleblueviolet, 
    size=small, 
    left=3mm, 
    right=3mm
  },
  title=#1,
  fonttitle=\color{white}\bfseries,
  arc=2mm,
  boxrule=0.5pt,
  left=5pt,
  right=5pt,
  top=10pt,
  bottom=5pt,
  fontupper=\small,
  breakable
}

\newtcolorbox{graybox}{
  colback=gray!10,
  colframe=gray!10,
  boxrule=0pt,
  arc=1mm,
  left=1mm,
  right=1mm,
  top=1mm,
  bottom=1mm,
  enhanced,
}

\begin{document}

\twocolumn

\title{
AnalogCoder-Pro: Unifying Analog Circuit  \\ Generation and Optimization via Multi-modal LLMs
}
\date{}

\author{
    Yao Lai,         \quad
    Souradip Poddar, \quad
    Sungyoung Lee,   \quad
    Guojin Chen,     \quad
    Mengkang Hu,     \\
    Bei Yu,          \quad
    Ping Luo,        \quad
    David Z.~Pan \\
    \thanks{Yao Lai, Mengkang Hu, and Ping Luo are with the Department of Computer Science, The University of Hong Kong, Hong Kong SAR.}
    \thanks{Souradip Poddar, Sungyoung Lee, and David Z.~Pan are with the Department of Electrical and Computer Engineering, The University of Texas at Austin, United States.}
    \thanks{Guojin Chen and Bei Yu are with the Department of Computer Science and Engineering, The Chinese University of Hong Kong, Hong Kong SAR.}
}

\maketitle
\thispagestyle{plain}
\pagestyle{plain}

\begin{abstract}

Despite recent advances, analog front-end design still relies heavily on expert intuition and iterative simulations, which limits the potential for automation.
We present AnalogCoder-Pro, a multimodal large language model (LLM) framework that integrates generative and optimization techniques.
The framework features a multimodal diagnosis-and-repair feedback loop that uses simulation error messages and waveform images to autonomously correct design errors. It also builds a reusable circuit tool library by archiving successful designs as modular subcircuits, accelerating the development of complex systems. Furthermore, it enables end-to-end automation by generating circuit topologies from target specifications, extracting key parameters, and applying Bayesian optimization for device sizing.
On a curated benchmark suite covering 13 circuit types, AnalogCoder-Pro successfully designed 28 circuits and consistently outperformed existing LLM-based methods in figures of merit.
The codes will be released at \href{https://github.com/laiyao1/AnalogCoderPro}{github.com/laiyao1/AnalogCoderPro}.


\end{abstract}

\section{Introduction}

Large Language Models (LLMs)\cite{achiam2023gpt} have recently demonstrated strong potential for automating chip design workflows\cite{zhong2023llm4eda, pan2025survey, fang2025survey}. Remarkable progress has been made in digital circuit design, including RTL code generation and correction~\cite{blocklove2023chip, chang2023chipgpt, thakur2023verigen, thakur2023autochip, fu2023gpt4aigchip, tsai2023rtlfixer, pei2024betterv, yao2024hdldebugger, zhu2025codev, akyash2025rtl++, liu2024craftrtl, deng2025scalertl, liu2024rtlcoder, zhao2024codev, liu2023chipnemo}, script generation~\cite{wu2024chateda, liu2023chipnemo, wu2025divergent}, and EDA documentation question answering~\cite{pu2024customized}.
While digital design tasks often benefit from standardized representations and abundant datasets, analog circuit design presents a fundamentally different set of hurdles that limit the direct transfer of such automation advances.

\begin{table}[!t]
\centering
\caption{\textbf{LLM-based Methods for Analog Design}}
\begin{adjustbox}{width=\columnwidth}
\begin{threeparttable}
\begin{tabular}{@{}l|cccccc@{}}
\toprule
\multirow{2}{*}{\textbf{Work}} & \textbf{Multiple} & \textbf{Training-} & \textbf{MLLM} & \textbf{Circuit} & \textbf{Circuit} & \textbf{Open-} \\
& \textbf{Types}$^1$ & \textbf{Free} & \textbf{Debug}$^2$ & \textbf{Gen.} & \textbf{Opt.} & \textbf{Source} \\
\midrule
CktGNN~\cite{dong2023cktgnn} &  &  &  & $\bullet$ & $\bullet$ & $\bullet$ \\
LADAC~\cite{liu2024ladac} & $\bullet$ & $\bullet$ &  &  & $\bullet$ &  \\
ADO-LLM~\cite{yin2024ado} & $\bullet$ & $\bullet$ &  &  & $\bullet$ &  \\
LaMAGIC~\cite{chang2024lamagic, chang2025lamagic2} &  &  &  & $\bullet$ & $\bullet$ &  \\
AnalogCoder~\cite{lai2025analogcoder} & $\bullet$ & $\bullet$ & $\circ$ & $\bullet$ &  & $\bullet$ \\
SPICEPilot~\cite{vungarala2024spicepilot} & $\bullet$ & $\bullet$ & $\circ$ & $\bullet$ &  &  \\
LEDRO~\cite{kochar2024ledro} &  & $\bullet$ &  &  & $\bullet$ &  \\
Aritsan~\cite{chen2024artisan} &  &  &  & $\bullet$ & $\bullet$ &  \\
AmpAgent~\cite{liu2024ampagent} &  & $\bullet$ &  &  & $\bullet$ &  \\
Atelier~\cite{shen2024atelier} &  & $\bullet$ &  & $\bullet$ & $\bullet$ &  \\
AnalogXpert~\cite{zhang2024analogxpert} &  & $\bullet$ & $\circ$  & $\bullet$ &  &  \\
Malasa-Chai~\cite{bhandari2025masalachai} & $\bullet$ &  &  & $\bullet$ &  & $\bullet$ \\
AnalogGenie/Lite~\cite{gao2025analoggenie, gao2025analoggenielite} & $\bullet$ &  &  & $\bullet$ & $\bullet$ & $\bullet$ \\
AnalogFed~\cite{li2025analogfedfederateddiscoveryanalog} & $\bullet$ &  &  & $\bullet$ & $\bullet$ &  \\
AutoCircuit-RL~\cite{vijayaraghavan2025autocircuit} & &  &  & $\bullet$ & $\bullet$ &  \\
SPICEAssistant~\cite{nau2025spiceassistant} &  & $\bullet$ & $\circ$ & $\bullet$ & $\bullet$ &  \\
\textbf{AnalogCoder-Pro} & $\bullet$ & $\bullet$ & $\bullet$ & $\bullet$ & $\bullet$ & $\bullet$ \\
\bottomrule
\end{tabular}
\begin{tablenotes}
\item $^1$ Whether the work supports multiple circuit type designs.
\item $^2$ $\bullet$ - Full multi-modal debugging capability by signal images (e.g., waveform, frequency response), $\circ$ - Text-only debugging.
\end{tablenotes}
\end{threeparttable}
\end{adjustbox}
\label{tab:model_comparison}
\end{table}

Analog circuit design automation, especially when applying LLMs, faces several key challenges, summarized as follows.
\textbf{(1) Complexity.} Analog circuit design is inherently complex due to diverse device behaviors and multidimensional performance trade-offs~\cite{razavi2000design,stehr2003performance,mueller2009successive, liu2007statistical}. Design flows rely heavily on expert heuristics and extensive simulations, since even minor adjustments in topology or device sizing can drastically alter circuit functionality, leading to a combinatorial explosion in the search space.
\textbf{(2) Abstraction gap.}
Digital design languages such as Verilog~\cite{thomas2008verilog} provide high-level functional abstractions, enabling concise descriptions (e.g., a digital adder can be specified in a single line of code). In contrast, analog design inherently operates at the device level, where even basic functions such as addition require explicitly wiring together multiple transistors and passive components. This abstraction gap limits the transfer of automation advances from digital to analog design.
\textbf{(3) Data scarcity.} Public code repositories contain very little hardware description language content—Verilog, for instance, accounts for less than 0.1\% of GitHub projects~\cite{guo2024deepseek}. SPICE~\cite{vladimirescu1994spice}, the predominant analog design language, has an even smaller footprint, making it particularly challenging for LLMs to acquire analog design rules effectively.
\textbf{(4) Underuse of multimodal information.} Simulation outputs such as waveforms and frequency responses contain critical information for analog circuit analysis. Yet, these non-textual modalities are rarely integrated into automated processing pipelines. This underutilization limits the ability of LLMs to diagnose issues and refine designs effectively.
\textbf{(5) Lack of unified topology–sizing optimization.} Topology generation and device sizing are typically conducted as separate, sequential stages, which can lead to suboptimal results. When inherent topological constraints cannot be addressed through parameter tuning, costly redesign cycles often follow.

Conventional analog automation tools provide valuable solutions but still face limitations.
Modern device sizing algorithms~\cite{wang2020gcn, settaluri2020autockt, budak2021dnn, budak2023apostle, choi2023reinforcement, li2021circuit, oh2024cronus, poddar2024insight, liu2009analog, yu2014advanced} improve sample efficiency yet are mostly restricted to predefined architectures. 
Topology exploration methods face high simulation costs~\cite{mcconaghy2007simultaneous,palmers2009massively}, depend on expert-crafted equations~\cite{veselinovic1995flexible,mcconaghy2008automated,zhao2020automated}, offer limited diversity~\cite{maulik1995integer}, and carry the risk of generating invalid structures~\cite{lu2023high}. 
In addition, automated synthesis approaches such as~\cite{poddar2024data} still demand substantial setup effort, limiting the applicability to predefined libraries.

Recent LLM-based approaches for analog circuit design, partly shown in Table~\ref{tab:model_comparison}, have emerged in three main directions, 
including topology generation via code or netlist synthesis~\cite{lai2025analogcoder, vungarala2024spicepilot, bhandari2025masalachai, zhang2024analogxpert}, LLM-assisted circuit sizing~\cite{chen2024artisan, shen2024atelier, liu2024ladac, yin2024ado, liu2024ampagent}, and question–answering assistants~\cite{abbineni2025muallmmultimodallargelanguage, zhao2025mmcircuiteval, shi2025amsbench, chen2025analogseeker}. 
While promising, these methods lack comprehensive solutions that integrate multimodal feedback and unified topology–sizing optimization. 
For example, AnalogCoder~\cite{lai2025analogcoder} and AnalogXpert~\cite{zhang2024analogxpert} have not yet incorporated waveform images for circuit analysis.
AnalogGenie~\cite{gao2025analoggenie} supports end-to-end topology generation through pre-training and fine-tuning, yet still depends on manual data collection and per-class fine-tuning, which restricts scalability.

To overcome these limitations, we propose \textbf{AnalogCoder-Pro}, a training-free and end-to-end multimodal LLM framework that jointly performs topology generation and device sizing via Python code generation. To the best of our knowledge, AnalogCoder-Pro is the first framework in this direction. Given a natural-language description of the target circuit’s functionality, structure, and performance goals, the framework autonomously generates, verifies, and optimizes analog circuits. For instance, when asked to ``Design a multi-stage op-amp that optimizes the FoM (GBW/Power),'' it can produce and iteratively refine a functional design. The key innovations of AnalogCoder-Pro include 
domain-specific design prompts, 
a feedback-enhanced flow that leverages multimodal inputs such as waveforms, 
a dedicated circuit tool library that enables flexible reuse of fundamental analog blocks, 
and LLM-assisted parameter extraction for circuit sizing. Together, these components allow the framework to deliver high-quality, verified analog front-end designs with minimal human intervention.

The main contributions of this work are:
\begin{itemize}
    \item Introduction of \textbf{AnalogCoder-Pro}, the first training-free multimodal LLM framework for end-to-end analog circuit design, capable of transforming natural-language design objectives into optimized netlists.
    \item Development of a feedback-enhanced multimodal design flow that leverages waveforms, simulation logs, and code diagnostics to iteratively refine designs and correct functional errors.
    \item Creation of a circuit tool library that enables flexible reuse and composition of fundamental analog blocks for complex circuit design.
    \item Proposal of an integrated parameter optimization framework that combines LLM-generated initial topologies with automated refinement for effective circuit sizing.
    \item Experimental validation showing that AnalogCoder-Pro outperforms existing LLM-based methods in both design success rate and circuit performance across diverse analog circuit types.
\end{itemize}

\section{Preliminaries}
\label{preliminaries}

\begin{figure}[!t]
\centering
\includegraphics[width=\columnwidth]{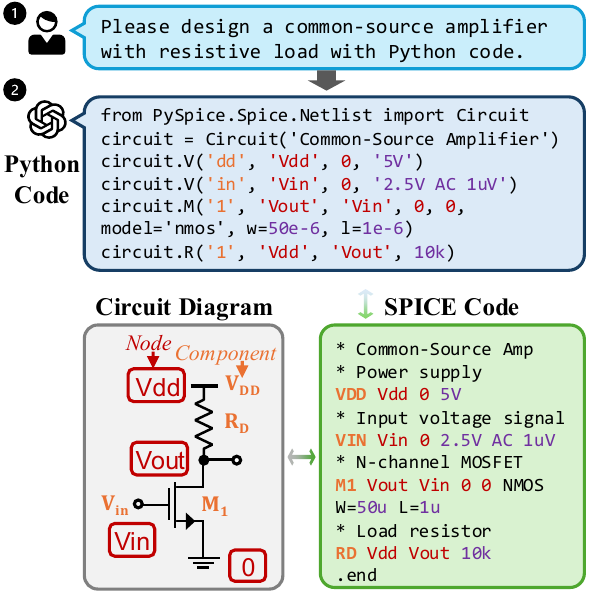}
\caption{\textbf{Code-based Analog Circuit Representation.} 
Given a design task, the LLM generates Python code with the PySpice library. 
The circuit can be equivalently represented as (1) a circuit diagram, (2) SPICE code, or (3) PySpice-based Python.}
\label{circuit_example}
\end{figure}

\subsection{Analog Circuit Design Challenges and Representation}

\textbf{Design Challenges.}
Analog circuit design is one of the most intricate stages in the integrated circuit flow, involving both front-end and back-end tasks. In the front-end, designers must synthesize suitable topologies and optimize device parameters to meet stringent, often conflicting performance criteria such as gain, bandwidth, and power consumption. Achieving a high figure of merit (FoM) requires navigating large, high-dimensional design spaces and managing complex trade-offs, which are further complicated by the strong coupling between topology and sizing decisions, where an optimal choice in one aspect may be suboptimal in another. The back-end translates these logical designs into manufacturable layouts~\cite{chen2020magical, chen2024llm, mangalagiri2024cdls, hammoud2024human}, ensuring process compliance.

Recent advances in artificial intelligence have brought new tools to analog design, including reinforcement learning~\cite{zhao2022deep, budak2021dnn, wang2020gcn}, graph-based generative models~\cite{dong2023cktgnn, zhao2019graph}, and language model-based approaches~\cite{lai2025analogcoder, poddar2024insight, zhang2024analogxpert, vungarala2024spicepilot, yin2024ado, kochar2024ledro, liu2024ladac, liu2024ampagent, bhandari2025masalachai}. Nevertheless, many of these techniques tackle topology generation and parameter optimization in isolation, which can result in incompatibilities that hinder overall performance and necessitate iterative redesigns.

To mitigate these issues, recent research has explored unified frameworks that jointly consider topology and sizing. AnalogGenie~\cite{gao2025analoggenie, gao2025analoggenielite} first pre-trains on a manually collected set of circuit topologies, then fine-tunes the model specifically on high-performance topologies. The method in~\cite{poddar2024data} automatically ranks candidate topologies according to user requirements, then performs sizing and verification on each, but remains limited to a predefined set of topologies. Artisan~\cite{chen2024artisan} focuses on behavioral-level synthesis for op-amps design, omitting device-level refinement. Thus, despite these advances, a truly scalable, general-purpose methodology for end-to-end, performance-driven analog design automation remains elusive.

\textbf{Circuit Representation.} SPICE~\cite{vladimirescu1994spice} is the industry-standard netlist format for circuit-level modeling and simulation, in which circuits are described as interconnected elements (e.g., resistors, voltage sources, MOSFETs) with associated parameters, as illustrated by the amplifier example in Fig.~\ref{circuit_example}.
In our framework, we integrate PySpice~\cite{salvaire2021pyspice}, a Python interface to SPICE simulators (e.g., Ngspice) that enables circuit construction, simulation control, and results parsing within a unified scripting environment. This approach leverages Python’s programmability and ecosystem for efficient circuit generation, simulation, and analysis, and aligns with the strong capability of LLMs in Python code generation~\cite{zheng2023codegeex}. Consequently, PySpice serves as a natural entry point for LLM-based circuit synthesis, replacing manual netlist creation.

\subsection{Multimodal Large Language Models}

Multimodal large language models (MLLMs) extend traditional LLMs by jointly processing visual and textual inputs. An MLLM typically includes a vision encoder and a text encoder, which produce feature tokens from images $I$ and text $T$:
\[
z_\text{vision} = E_\text{vision}(I), \quad z_\text{text} = E_\text{text}(T), \quad
y = \mathrm{LM}([z_\text{vision}, z_\text{text}]),
\]
where $y$ is the output sequence.

\begin{figure}[!ht]
\centering
\includegraphics[width=0.9\columnwidth]{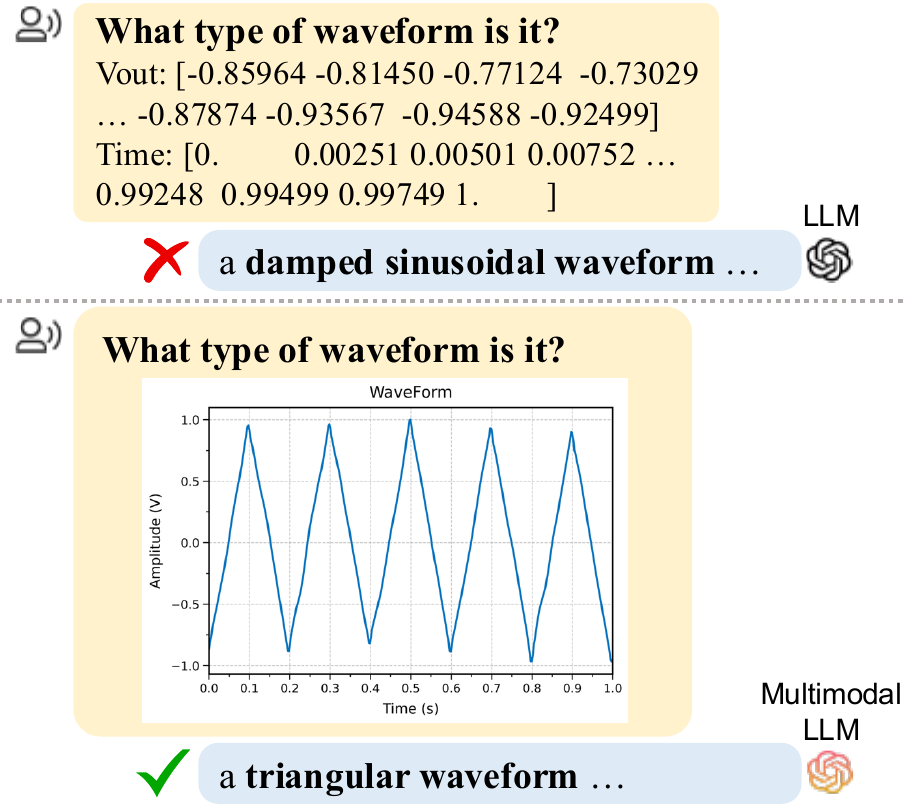}
\caption{\textbf{LLMs vs. MLLMs in Waveform Identification.} LLMs provided with textual data (top) misidentify a noisy triangular waveform, whereas MLLMs, given the corresponding waveform image (bottom), correctly identify its type.}
\label{example_mllm}
\end{figure}

In EDA, MLLMs enable richer interpretation of circuit diagrams and waveform plots, outperforming text-only models in pattern recognition tasks. For example, Fig.~\ref{example_mllm} shows that a noisy triangular waveform is misclassified by an LLM (text input only) as a damped sinusoid, whereas an MLLM correctly identifies it from the waveform image. This visual reasoning capability is crucial for reliable waveform identification, thereby improving circuit debugging and validation.

\section{Framework and Methods}
\label{framework}

\subsection{Framework Overview}
\begin{figure*}[!t]
\centering
\includegraphics[width=\textwidth]{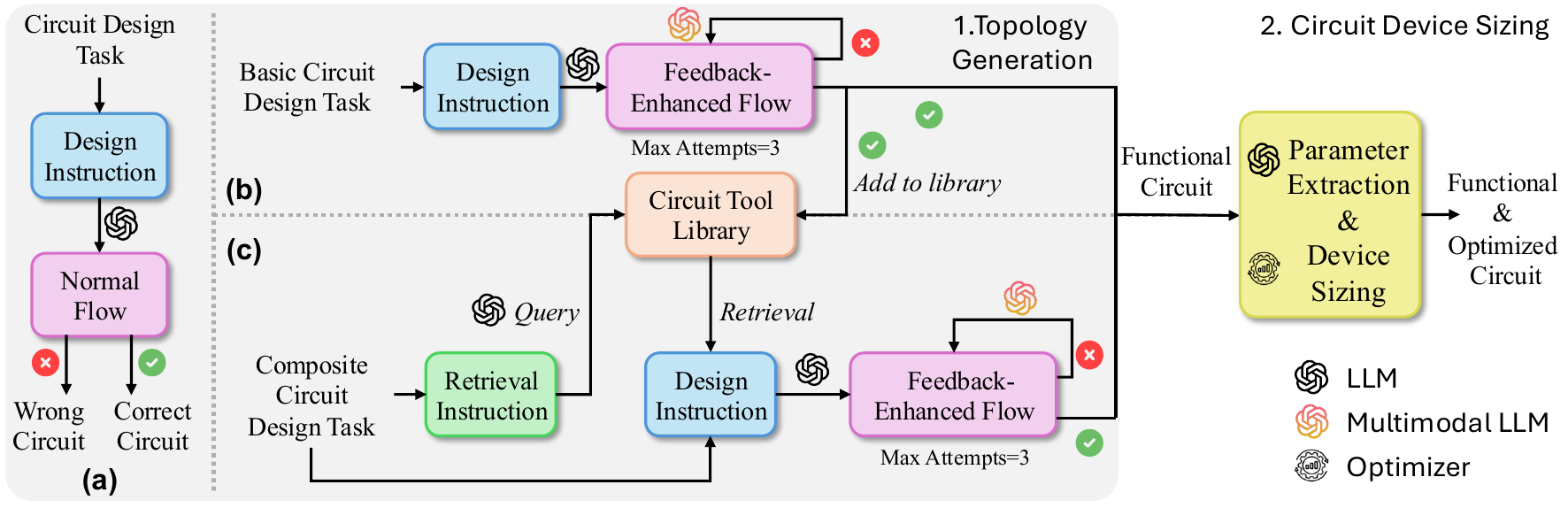}
\caption{{\textbf{Framework Overview.} 
AnalogCoder-Pro consists of two main stages: (1) topology generation and (2) circuit device sizing. 
(a) \textbf{Standard method.} A direct design flow often results in failed circuits.  
(b) \textbf{AnalogCoder-Pro for basic circuits.} design prompts are processed through a multimodal feedback‑enhanced flow, where LLMs iteratively correct errors; successful circuits are stored in the circuit tool library.  
(c) \textbf{AnalogCoder-Pro for composite circuits.} Relevant subcircuits are retrieved from the library and incorporated into the design prompt to facilitate composite circuit generation. 
After topology generation, the device sizing is conducted, which gives the final functional and optimized circuit.
}}
\label{method_overview}
\end{figure*}

AnalogCoder-Pro is a multimodal LLM-based framework that interprets natural-language task descriptions to automatically generate Python code representing functional, optimized analog circuits.
As shown in Fig.~\ref{method_overview}, it consists of multiple modules, including prompt engineering, a multimodal feedback‑enhanced flow, and a circuit tool library, followed by device sizing.
Prompt engineering enhances the design reasoning through strategic, problem‑solving prompts.
The multimodal feedback‑enhanced flow incorporates multiple checks, including waveform validation, to feed error messages back to the LLM, enabling it to correct failed designs.
The circuit tool library serves as a modular repository that organizes designed circuits into reusable components, enabling efficient retrieval and integration into more complex designs.
Following topology generation, it conducts device sizing, including parameter extraction and optimization, to finalize the circuit design.

\subsection{Prompt Design}
To leverage the code generation capabilities of LLMs for circuit design, we adopt a systematic prompt design strategy, comprising: (1) \textit{programming language selection}, 
(2) \textit{in-context learning}~\cite{dong2022survey}, 
and (3) \textit{Chain-of-Thought reasoning}~\cite{wei2022chain}.
Given the superior performance of LLMs in Python code generation, owing to the prevalence of Python in model pre‑training~\cite{zheng2023codegeex}, we instruct the models to produce executable Python code compatible with the PySpice library, thereby overcoming the lack of SPICE code in common training datasets~\cite{guo2024deepseek}.
For in-context learning, we provide a single detailed example of a two-stage amplifier, enabling the LLM to generalize and standardize its output across diverse design tasks. 
To further enhance design quality, we employ Chain-of-Thought prompting~\cite{wei2022chain}, which guides the LLM to generate a step‑by‑step design plan before producing the corresponding code.
\looseness=-1

\begin{figure*}[!ht]
\centering
\includegraphics[width=\textwidth]{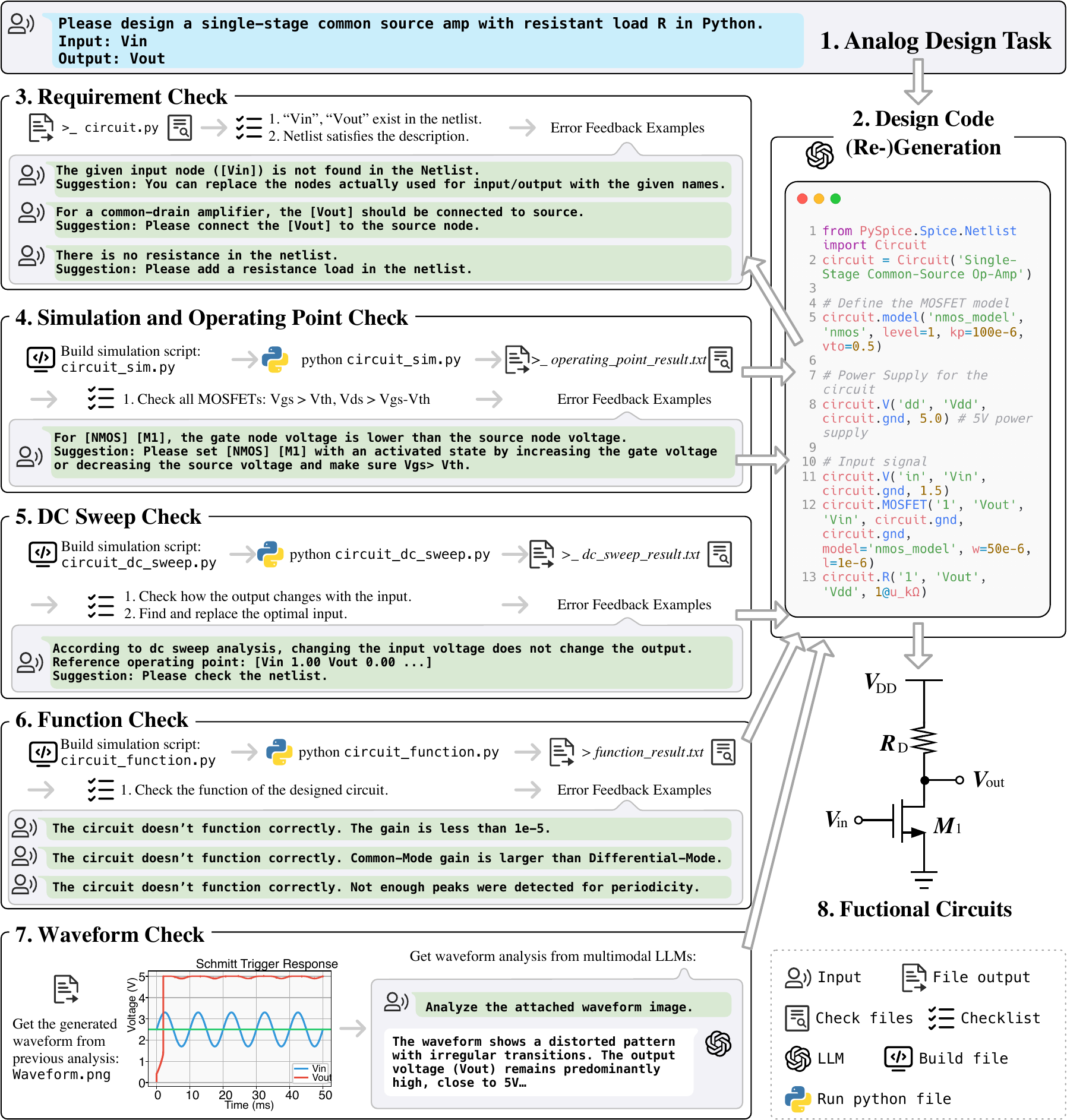}
\caption{
    \textbf{Multimodal Feedback-Enhanced Design Flow.}
    A feedback-enhanced flow in which the LLM integrates textual error messages and waveform anomalies to iteratively refine designs without human intervention. Error messages are generated from predefined templates, ensuring applicability across diverse circuit types. Waveform images are automatically produced from simulation data using the matplotlib package. For illustration, the waveform example is derived from a Schmitt trigger circuit.
}
\label{overall}
\end{figure*}

\subsection{Multimodal Feedback-enhanced Flow}
To ensure the correctness of analog circuit designs, the verification in feedback flow is divided into five stages as shown in Fig.~\ref{overall}: (1) \textit{Requirement Check}, which ensures basic design elements such as the input voltage ($V_{in}$) and output voltage ($V_{out}$) are present, and validates the inclusion of essential components; (2) \textit{Simulation and Operating Point Check}, where the circuit is simulated to detect issues like floating nodes and verify MOSFET operating conditions (e.g., $V_{gs} > V_{th}$); (3) \textit{DC Sweep Check}, which involves varying $V_{in}$ to observe changes in $V_{out}$, ensuring signal path integrity and identifying optimal bias points; (4) \textit{Function Check}, where simulations validate functionalities such as gain and periodicity; and (5) \textit{Waveform Check}: simulation data are plotted as waveforms (e.g., transient, AC sweep) using the Matplotlib package, analogous to an oscilloscope, to visually verify the output signal.
The used waveform types for different circuit types are shown in Table~\ref{tab:circuit_simulation}. 

If errors are detected during these stages, diagnostic messages (e.g., ``Missing $V_{\mathrm{in}}$ node'' or ``Gain is less than $1\text{e}{-5}$'') are generated from predefined templates to guide iterative refinement. 
The framework permits up to three design iterations, each informed by error feedback and waveform analysis, to progressively improve the design.

\begin{table}[!ht]
\centering
\caption{\textbf{Waveform Types for Different Circuit Types}}
\label{tab:circuit_simulation}
\begin{adjustbox}{width=\columnwidth}
\begin{tabular}{l|c|ccccc}
\toprule
\textbf{Circuit Type} & \textbf{Abbr.} & \textbf{Transient} & \textbf{AC} & \textbf{FFT} & \textbf{DC Sweep} & \textbf{DC Transfer} \\
\midrule
Mixer & Mix & $\bullet$ & & $\bullet$ & & \\
Comparator & Comp & & & & $\bullet$ & \\
Filter & Filt & & $\bullet$ & & & \\
Oscillator & Osc & $\bullet$ & & & & \\
Integrator & Int & $\bullet$ & & & & \\
Differentiator & Diff & $\bullet$ & & & & \\
Schmitt trigger & Sch & $\bullet$ & & & & $\bullet$ \\
\bottomrule
\end{tabular}
\end{adjustbox}
\end{table}

\subsection{Circuit Tool Library}
\begin{figure*}[!t]
  \centering
  \includegraphics[width=\textwidth]{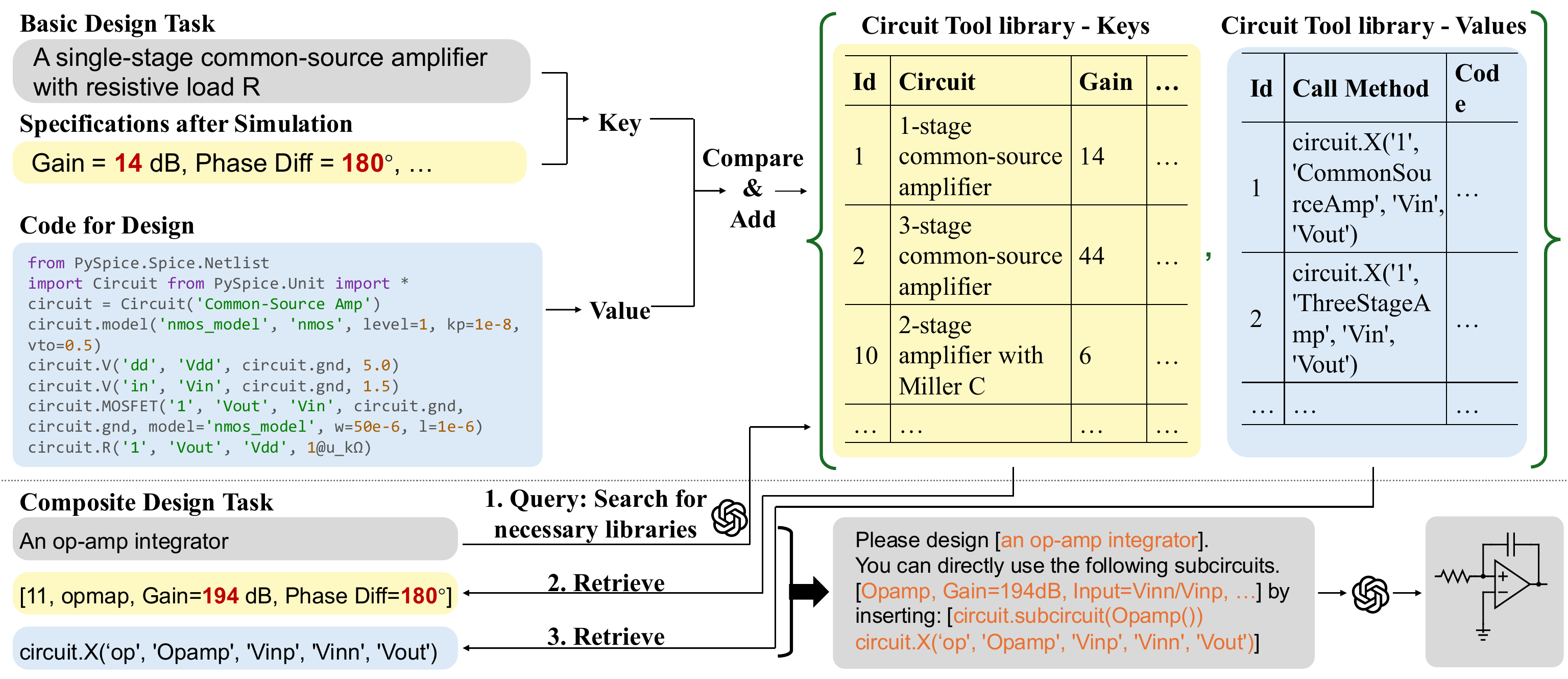}
  \caption{{\textbf{Circuit Tool library.} \textbf{Top:} Addition of new tools derived from successfully designed basic circuits. Here, descriptions and specifications are keys, while design codes are stored as values.
\textbf{Bottom:} Retrieval of tools from the library for designing composite circuits. The process begins with the LLM querying the necessary tools using the task description. Subsequently, the keys and values of the retrieved tools, with the task description, are employed as instructions for circuit design.
}}
  \label{skill_library}
\end{figure*}

As analog circuit design tasks grow more complex, it becomes increasingly challenging for LLMs to generate correct circuits. 
To address this, basic circuits can be modularized into subcircuits, facilitating their integration into composite designs. 
Inspired by tool-based LLM studies~\cite{wang2023voyager, qin2023toolllm}, we developed a circuit tool library to store correctly designed subcircuits for reuse in more complex designs.
As illustrated in Fig.~\ref{skill_library}, our approach comprises two key processes: adding circuits to the library and retrieving circuits from the library for composite designs. 
After AnalogCoder-Pro completes a basic circuit design, the corresponding circuit representations and specifications (e.g., gain) are stored in the library. 
For repeated tasks, the library retains the optimal design based on specifications. 
Subcircuit descriptions and circuit information serve as query keys, while circuit representations and usage instructions serve as values. 
During composite circuit design, the task description is used to query the library, retrieving the required subcircuits along with their specifications.
The design prompt then integrates this information with the task description, enabling AnalogCoder-Pro to generate the final design. 
For example, when designing an op-amp integrator, the process involves retrieving the single-stage op-amp subcircuit and integrating it into the design.

\subsection{Topology Generation Example}

Below is a complete example conversation for circuit topology generation in AnalogCoder-Pro. 
In this example, the LLM is tasked with designing an RC-shift oscillator, with the op-amp subcircuit from the circuit tool library provided for reuse. 
In addition, the MLLM feedback-enhanced flow is introduced for circuit diagnosis and repair.

\begin{promptbox}[Instruction (Round 1; LLM)]
\footnotesize
You aim to design a topology for a given circuit described in the text. 
Please ensure your designed circuit topology works properly and achieves the design requirements.

To make the task easier, you can directly use the opamp subcircuits by inserting:
\begin{minted}
[
frame=none,
baselinestretch=1,
fontsize=\scriptsize,
]
{python}
from opamp import *
# Declare the operational amplifier subcircuit
circuit.subcircuit(Opamp())
# Create a subcircuit instance
# Parameter order: instance name, subcircuit name, 
# the non-inverting input, the inverting input,
# the output
circuit.X('1', 'Opamp', 'Vinp', 'Vinn', 'Vout')
\end{minted}

The DC bias voltage for both input terminals (Vinn and Vinp) is 2.5V. For AC coupling/grounding purposes, these terminals should be referenced to the 2.5V power supply rather than ground.
\vspace{0.5em}
Here is an example:

...\textit{(An example code for an opamp circuit with resistive load)}
\vspace{0.5em}

As you have seen, the output of your designed topology should be in a complete Python code, describing the topology of integrated analog circuits according to the design plan. 

Please make sure your Python code is compatible with PySpice. 

Please give the runnable code without any placeholders.

\vspace{0.5em}

There are some tips you should remember all the time:
\begin{itemize}
  \item For the MOSFET definition \texttt{circuit.MOSFET(name, drain, gate, source, bulk, model, w=w1, l=l1)}, be careful about the parameter sequence.
  \item ...
\end{itemize}

Please first give a detailed design plan and then write the code.

\mycenterheading{Question}

Design \textcolor{boxemph}{\textbf{an RC phase-shift oscillator}}.

Input node name: \textcolor{boxemph}{\textbf{-}}.

Output node name: \textcolor{boxemph}{\textbf{Vout}}.

\mycenterheading{Answer}

\end{promptbox}

\begin{promptbox}[Response (Round 1; LLM)]
\footnotesize

An RC phase-shift oscillator uses an amplifier with a feedback network consisting of three RC stages. Each RC stage provides approximately 60° phase shift, for a total of 180° through the feedback network. Combined with the 180° phase shift from the inverting amplifier, this creates a 360° phase shift around the loop, satisfying the phase criterion for oscillation.

\mysubheading{Design Requirements:}
\begin{itemize}
  \item Create an oscillator circuit using an operational amplifier.
  \item Implement three RC stages for phase shifting.
  \item Set appropriate gain to ensure oscillation.
  \item No input node is needed as this is an oscillator (self-generating signal).
  \item Output node should be named \texttt{Vout}.
\end{itemize}

\mysubheading{Circuit Components:}
\begin{itemize}
  \item Operational amplifier in inverting configuration.
  \item Three RC stages for phase shifting.
  \item Feedback resistors to set appropriate gain.
\end{itemize}

\mysubheading{Design Details:}
\begin{itemize}
  \item For an RC phase-shift oscillator to work, the amplifier gain must be at least $29$ (approximately $29.3$) to overcome losses in the RC network.
  \item Three identical RC stages will be used, with $R = 10\,\mathrm{k}\Omega$ and $C = 10\,\mathrm{nF}$.
  \item The oscillation frequency will be approximately $f = \frac{1}{2\pi RC\sqrt{6}} \approx 650\,\mathrm{Hz}$.
  \item The feedback resistor $R_f$ will be set to provide sufficient gain ($> 29$).
  \item The input resistor $R_{in}$ will be chosen to achieve this gain.
\end{itemize}

Let's implement this design:

\begin{minted}
[
frame=none,
baselinestretch=1,
fontsize=\scriptsize,
]
{python}
from PySpice.Spice.Netlist import Circuit
from PySpice.Unit import *
from opamp import *
circuit = Circuit('RC Phase-Shift Oscillator')
# Define the MOSFET models
circuit.model('nmos_model', 'nmos', 
level=1, kp=100e-6, vto=0.5)
circuit.model('pmos_model', 'pmos', 
level=1, kp=50e-6, vto=-0.5)
# Power supply
circuit.V('dd', 'Vdd', circuit.gnd, 5@u_V)
# Set reference voltage (2.5V) as virtual ground
circuit.V('ref', 'Vref', circuit.gnd, 2.5@u_V)
# Declare the opamp subcircuit
circuit.subcircuit(Opamp())
# Create opamp instance (inverting configuration)
circuit.X('op', 'Opamp', 'Vref', 'feedback', 'Vout')
# RC phase-shift network (three stages)
# First RC stage
circuit.R('1', 'Vout', 'node1', 10@u_kOhm)
circuit.C('1', 'node1', 'Vref', 10@u_nF)
# Second RC stage
circuit.R('2', 'node1', 'node2', 10@u_kOhm)
circuit.C('2', 'node2', 'Vref', 10@u_nF)
# Third RC stage
circuit.R('3', 'node2', 'feedback', 10@u_kOhm)
circuit.C('3', 'feedback', 'Vref', 10@u_nF)
# Feedback resistor and input resistor to set gain
# Gain needs to be > 29 for oscillation
# Feedback resistor
circuit.R('f', 'feedback', 'Vout', 300@u_kOhm)  
# Input resistor to set gain
circuit.R('in', 'feedback', 'Vref', 10@u_kOhm)  
simulator = circuit.simulator()
\end{minted}

\end{promptbox}


During functional testing, the oscillator failed to maintain stable oscillation—the amplitude gradually decreased to zero. The waveform, generated with Matplotlib, was submitted to a multimodal LLM for further analysis. We give an example as follows. The parts highlighted in \textcolor{orange}{\textbf{orange}} will be updated according to the circuit type.

\begin{promptboxmllm}[Instruction (Multimodal LLM)]

Analyze the attached waveform output image from \textcolor{orange}{\textbf{an RC phase-shift oscillator circuit.}} 

\textcolor{orange}{\textbf{Vout should exhibit periodic oscillations.}} The actual output waveform differs. 

\textcolor{orange}{\textbf{The testbench initializes the op-amp non-inverting input (Vinp) to 2.51V and the inverting input (Vinn) to 2.5V.}}

\begin{center}
    \includegraphics[width=0.8\columnwidth]{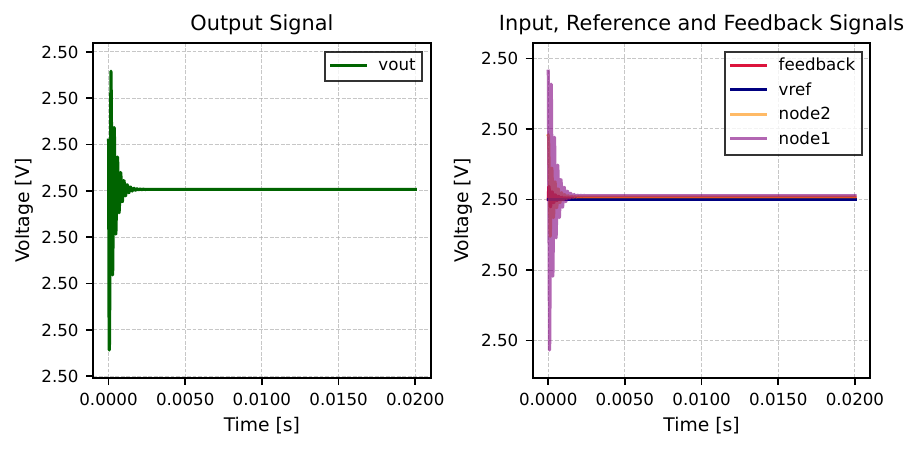}
\end{center}
\end{promptboxmllm}

\begin{promptboxmllm}[Response (Multimodal LLM)]
\begin{graybox}
The waveform shows a damped oscillation that quickly settles to a steady DC level rather than exhibiting the expected sustained periodic oscillations for an RC phase-shift oscillator. 

The output signal (vout) initially displays several rapidly decaying oscillations with a peak amplitude deviation of approximately ±0.01V around the 2.5V level, occurring within the first 0.001 seconds of simulation time.
\end{graybox}
\end{promptboxmllm}

The waveform analysis from the MLLM and the error messages obtained during the check process are jointly fed back to the LLM, enabling it to revise the design and regenerate a correct circuit implementation.

\begin{promptbox}[Instruction (Round 2; LLM)]
I am encountering an error when running the PySpice code. Below is the error message:

\begin{itshape}
Average oscillation period: 0.000002 s

Maximum amplitude: 0.000000 V

The oscillation amplitude is too small.
\end{itshape}

\mysubheading{Waveform Analysis:}

\begin{graybox}
The waveform shows a dumped oscillation \ldots{} \textit{(the output of MLLM waveform analysis)}

\end{graybox}

Identify any structural errors in this circuit's topology (e.g., missing components, incorrect connections) and rewrite the complete corrected code from scratch, with each modification marked and explained in comments.

\end{promptbox}

\begin{promptbox}[Response (Round 2; LLM)]
\footnotesize
\mysubheading{Corrected RC Phase-Shift Oscillator Design}

The main issues with the previous design were:

\begin{itemize}
  \item Insufficient gain to sustain oscillations
  \item Improper RC network configuration
  \item Lack of initial condition to start oscillations
\end{itemize}

Let me redesign the circuit with proper topology and gain:

\begin{minted}
[
frame=none,
baselinestretch=1,
fontsize=\scriptsize,
]
{python}
from PySpice.Spice.Netlist import Circuit

# ... (defining MOSFET models and power supply is omitted)

# Declare the opamp subcircuit
circuit.subcircuit(Opamp())
# Create opamp instance (inverting configuration)
# CORRECTION: Using proper non-inverting input 
# and inverting input connections
circuit.X('op', 'Opamp', 'Vref', 'feedback', 'Vout')
# CORRECTION: Proper RC ladder network 
# for phase-shift oscillator
# First RC stage - from output to first node
circuit.R('1', 'Vout', 'node1', 10@u_kOhm)
circuit.C('1', 'node1', 'Vref', 10@u_nF)
# Second RC stage
circuit.R('2', 'node1', 'node2', 10@u_kOhm)
circuit.C('2', 'node2', 'Vref', 10@u_nF)
# Third RC stage
circuit.R('3', 'node2', 'feedback', 10@u_kOhm)
circuit.C('3', 'feedback', 'Vref', 10@u_nF)
# CORRECTION: Significantly increased 
# gain to ensure oscillation. For RC phase-shift 
# oscillator, gain must be at least 29
# Using much higher gain (100) 
# to ensure strong oscillation
# Feedback resistor (increased)
circuit.R('f', 'feedback', 'Vout', 1000@u_kOhm)  
# Input resistor
circuit.R('in', 'feedback', 'Vref', 10@u_kOhm)   
# CORRECTION: Add a small initial condition 
# to kickstart oscillation
# This is needed because SPICE simulations 
# start from a steady state
circuit.C('kickstart', 'Vout', 'Vref', 1@u_nF, 
initial_condition=0.1@u_V)
simulator = circuit.simulator()
\end{minted}

\mysubheading{Key corrections:}

\begin{itemize}
  \item \textbf{Increased gain}: The gain of the amplifier is now much higher ($R_f/R_{in} = 1000\text{k}/10\text{k} = 100$), well above the minimum required gain of 29 for an RC phase-shift oscillator.
  \item \textbf{Proper RC network}: The RC network is configured correctly with each capacitor connected to the reference voltage.
  \item \textbf{Initial condition}: Added a capacitor with an initial condition to kickstart the oscillation.
  \item \textbf{Feedback path}: Ensured proper feedback from the last RC stage to the inverting input of the op-amp.
\end{itemize}

This circuit should now produce sustained oscillations at approximately 650 Hz.
\end{promptbox}

\begin{figure}[!h]
\centering
\includegraphics[width=0.8\columnwidth]{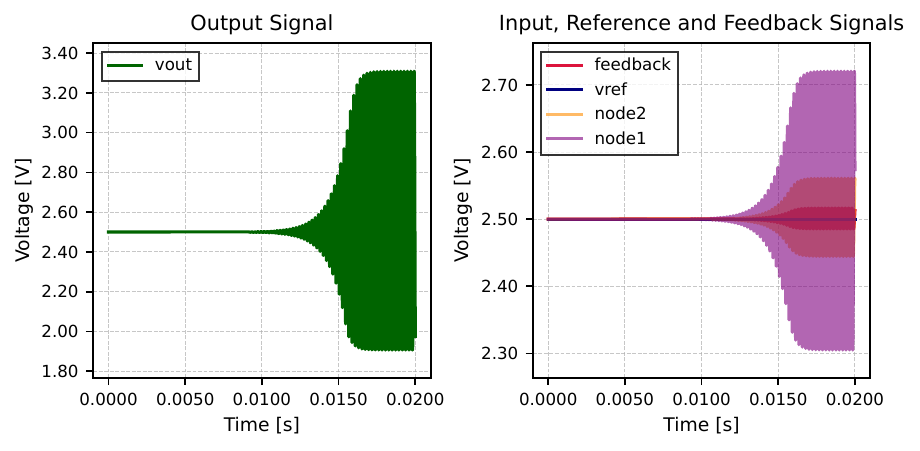}
\caption{\textbf{Waveform of a functional RC Phase-Shift Oscillator}}
\label{waveform_oscillator}
\end{figure}

Based on the analysis of the waveforms, the oscillator was revised primarily by adjusting the feedback resistor value to increase the loop gain. This modification enables the circuit to sustain stable oscillations. The post-simulation waveform of the revised oscillator is shown in Fig.~\ref{waveform_oscillator}.

\subsection{LLM-assisted Circuit Device Sizing}

Upon completion of circuit topology generation, the resulting code contains nominal parameters (e.g., transistor sizing, bias voltages, capacitances, resistances) that ensure functionality but may be suboptimal in performance. We address this by instructing the LLM to:
(1) analyze the generated netlist,
(2) extract all tunable parameters with appropriate search ranges, and
(3) reformat the circuit into a parameterized template for use with optimization frameworks (e.g., Optuna).
We give an example instruction and response as follows. 
In this case, the LLM extracts the load resistance value and the transistor width as tunable parameters, producing a parameterized circuit implementation that can be directly invoked by an optimization framework.

\begin{promptbox}[Instruction (LLM)]
\footnotesize
I need to convert my PySpice circuit netlist into a parameterized form for automated optimization. Please transform the following original circuit description into a parameter-based implementation with appropriate search ranges.

For reference, here is an example of what an original circuit might look like: \ldots

Here is my actual circuit code:
\begin{minted}[
frame=none, baselinestretch=1, fontsize=\scriptsize
]{python}
circuit = Circuit('Single-Stage Amplifier')
# MOSFET model
circuit.model('nmos_model', 'nmos', **nmos_params)
# ...
circuit.MOSFET('1', 'Vout', 'Vin', circuit.gnd, 
    circuit.gnd, model='nmos_model', w=5e-6, l=0.045e-6)
circuit.R('1', 'Vout', 'Vdd', 10@u_kOhm)
\end{minted}
\mysubheading{Required Output}
\begin{enumerate}
    \item A circuit creation function that uses a \texttt{params} dictionary.
    \item A parameter search range dictionary (for Optuna or similar optimization tools).
    \item Reasonable parameter constraints with tailored search ranges:
    \begin{itemize}
        \item Transistor width (\(W\)) should be within \(1\text{--}500\times\) the corresponding length (\(L\)).
        \item \ldots
    \end{itemize}
\end{enumerate}
\end{promptbox}

\begin{promptbox}[Response (LLM)]
\begin{minted}[
frame=none, baselinestretch=1, fontsize=\scriptsize
]{python}
def create_circuit(params):
    circuit = Circuit('Single-Stage Amplifier')
    circuit.model('nmos_model', 'nmos', **nmos_params)
    # ...
    circuit.MOSFET('1', 'Vout', 'Vin', circuit.gnd, 
                   circuit.gnd, model='nmos_model',
                   w=params['w_M1'], l=0.045e-6)
    circuit.R('1', 'Vout', 'Vdd', f"{params['r_load']}k")
    return circuit

param_ranges_definition = {
  'w_M1': {'min': 0.045e-6, 'max': 22.5e-6, 'log': True},
  'r_load': {'min': 5, 'max': 20, 'log': True},
}
initial_params = {'w_M1': 5e-6, 'r_load': 10}
\end{minted}
\end{promptbox}

\begin{figure}[!t]
\centering
\includegraphics[width=\columnwidth]{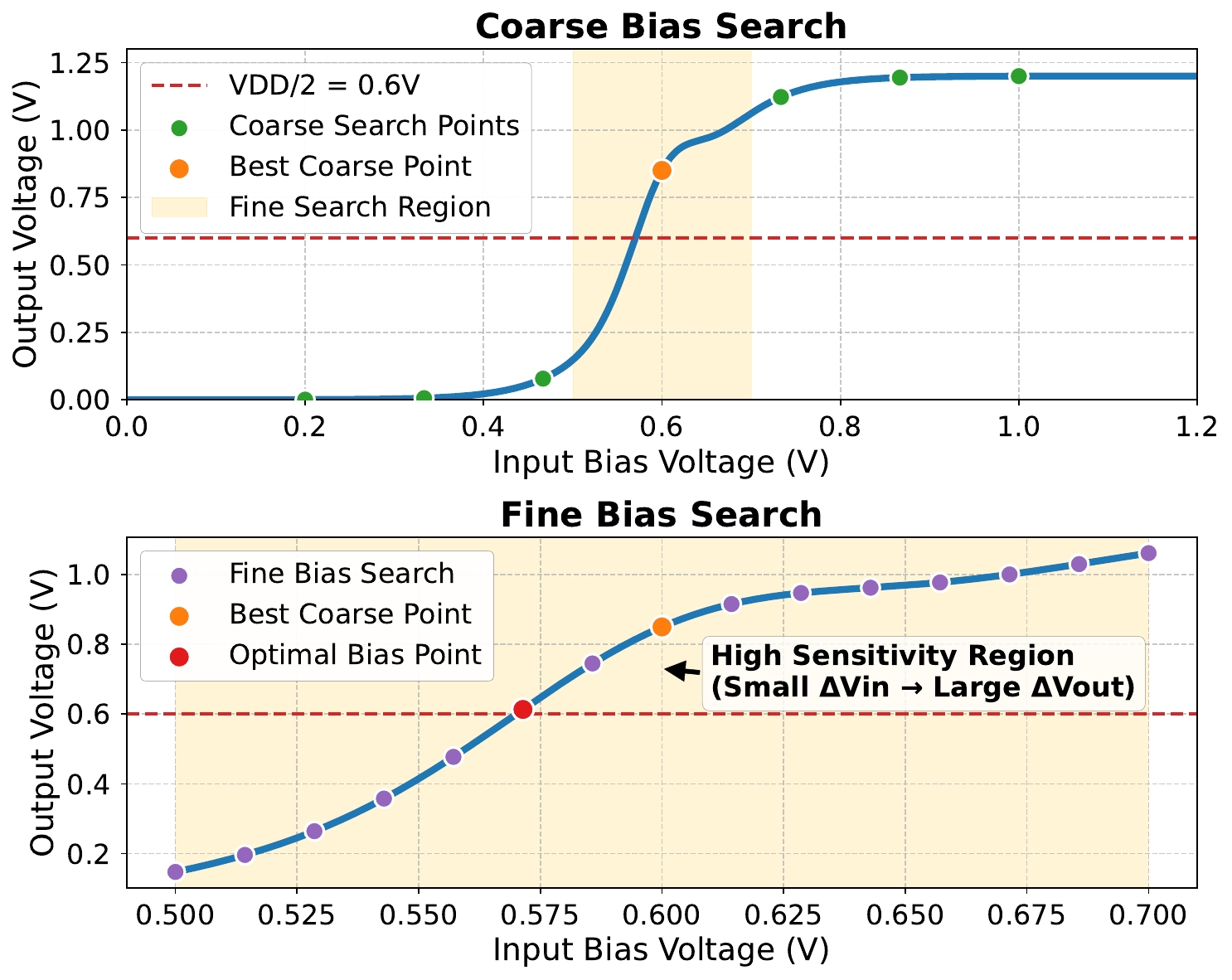}
\caption{\textbf{Multi-Resolution Search for Input Bias.}
It conducts a coarse-to-fine DC sweep to automatically locate the input bias that sets the output voltage closest to $V_\mathrm{DD}/2$, ensuring proper circuit biasing while reducing optimization dimensionality.
}
\label{transfer_curve}
\end{figure}

With the tunable parameters, we use the Bayesian optimizer for circuit sizing. 
For each parameter set proposed by the optimizer, the input voltage bias is not treated as an optimization variable, because the output voltage is highly sensitive to the bias point: even small changes in input voltage can cause large variations in output voltage.
Therefore, we propose a novel multi-resolution input bias search process to determine the optimal input bias (\(V_{in}\)) before the main optimization loop. 
As illustrated in Fig.~\ref{transfer_curve}, this algorithm performs DC sweeps at progressively refined resolutions to locate the bias point that sets the output voltage close to \(V_{DD}/2\) in single-ended, single-supply configurations, providing a balanced operating point and placing the key transistors in their intended operating regions. Once this optimal bias is established, we proceed with a comprehensive circuit simulation to evaluate the figure of merit (FoM), which then provides the reward signal for the optimizer. This decoupling strategy reduces the dimensionality of the optimization space, improves convergence efficiency, and ensures proper circuit biasing for each parameter configuration under evaluation.

\section{Experimental Results}
\label{experimental}

\subsection{Experimental settings}

\begin{figure}[!th]
\centering
\includegraphics[width=\columnwidth]{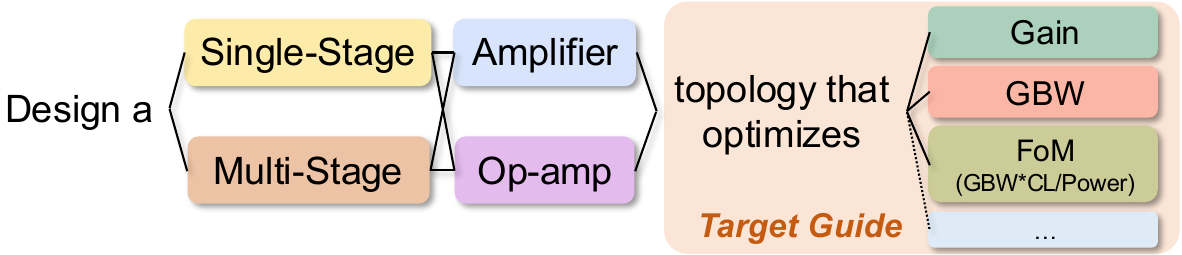}
\caption{
\textbf{Unified Generation and Optimization Prompt Strategies.}
All prompt variants illustrated here are evaluated in Table~\ref{tab:comparison}, with the target-guided objective (e.g., Gain, GBW, or user-defined FoM) being optional for topology optimization.
}
\label{unified_design_task}
\end{figure}

\textbf{Benchmark.} 
For topology generation, we consider 28 design tasks spanning 13 circuit types, including amplifiers, operational amplifiers, current mirrors, mixers, filters, and comparators, as summarized in Table~\ref{benchmark}. Each circuit type is accompanied by functional verification code that validates the generated netlists through simulation results. 
The tasks are categorized into three difficulty levels (easy, medium, and hard) according to the number of components and the complexity of their interconnections. 
A circuit is considered composite when it involves the reuse of subcircuits.
In our task design, Tasks 9 and 22–28 are composite circuits that utilize the circuit tool library, whereas all other tasks correspond to basic circuits.
For circuit device sizing, we evaluated the ability of LLMs to generate performance-driven analog circuit topologies using 12 distinct design instructions (see Fig.~\ref{unified_design_task}).

\textbf{Metrics.}
For topology generation, we adopt the widely-used $\text{Pass@}k$, which measures the probability that at least one correct solution appears in $k$ sampled generations. 
Formally, it is defined as
$\text{Pass@}k = 1 - C_{n-c}^k / C_n^k$,
where $n$ denotes the total number of attempts, $c$ denotes the number of correct generations, 
and $C_n^k$ denotes the binomial coefficient.
In our experiments, we set $n=30$.
Note that we use the feedback-enhanced flow, allowing each generation task to be revised up to two additional times based on error information.
For circuit device sizing, we align with AnalogGenie~\cite{gao2025analoggenie}, using $\text{FoM}=\text{GBW}\times C_L/\text{Power}$ as the primary optimization target.
For completeness, we also evaluate Gain and GBW as alternative optimization goals.

\begin{table*}[!th]
\caption{\textbf{Benchmark Descriptions.} Difficulties are distinguished by background colors (\textcolor{easy_green}{\textbf{easy}}, \textcolor{medium_orange}{\textbf{medium}}, and \textcolor{hard_red}{\textbf{hard}}).}
\centering
\label{benchmark}
\begin{adjustbox}{max width=\textwidth}
\begin{tabular}{r|l|l|r|l|l}
\toprule
\textbf{ID} & \textbf{Type} & \textbf{Circuit Description} & \textbf{ID} & \textbf{Type} & \textbf{Circuit Description} \\
\midrule
\rowcolor{lightGreen} 
1  & Amplifier & Common-source amp. with R load & \cellcolor{lightOrange}15 & \cellcolor{lightOrange}Amplifier & \cellcolor{lightOrange}2-stage amplifier with Miller compensation \\
\rowcolor{lightGreen} 
2  & Amplifier & 3-stage common-source amplifer with R loads & \cellcolor{lightOrange}16 & \cellcolor{lightOrange}Current Mirror & \cellcolor{lightOrange}Cascode current mirror \\
\rowcolor{lightGreen} 
3  & Amplifier & Common-drain amp. with R load & \cellcolor{lightOrange}17 & \cellcolor{lightOrange}Opamp & \cellcolor{lightOrange}Op-amp with active current mirror loads \\
\rowcolor{lightGreen} 
4  & Amplifier & Common-gate amp. with R load & \cellcolor{lightOrange}18 & \cellcolor{lightOrange}Opamp & \cellcolor{lightOrange}Common-source op-amp with R loads \\
\rowcolor{lightGreen} 
5  & Amplifier & Cascode amp. with R load & \cellcolor{lightOrange}19 & \cellcolor{lightOrange}Mixer & \cellcolor{lightOrange}A Gilbert cell mixer \\
\rowcolor{lightGreen} 
6  & Inverter  & NMOS inverter with R load & \cellcolor{lightRed}20 & \cellcolor{lightRed}Opamp & \cellcolor{lightRed}Cascode op-amp with cascode loads \\
\rowcolor{lightGreen} 
7  & Inverter  & Logical inverter & \cellcolor{lightRed}21 & \cellcolor{lightRed}Opamp & \cellcolor{lightRed}2-stage op-map with active loads \\
\rowcolor{lightGreen} 
8  & Current Mirror & NMOS constant current source with R load & \cellcolor{lightRed}22 & \cellcolor{lightRed}Oscillator & \cellcolor{lightRed}Wien Bridge oscillator \\
\rowcolor{lightGreen} 
9  & Comparator & an op-amp comparator & \cellcolor{lightRed}23 & \cellcolor{lightRed}Oscillator & \cellcolor{lightRed}RC Shift oscillator \\
\rowcolor{lightGreen} 
10 & Filter & a passive low-pass filter & \cellcolor{lightRed}24 & \cellcolor{lightRed}Integrator & \cellcolor{lightRed}Op-map integrator \\
\rowcolor{lightGreen} 
11 & Filter & a passive high-pass filter & \cellcolor{lightRed}25 & \cellcolor{lightRed}Differentiator & \cellcolor{lightRed}Op-map differentiator \\
\rowcolor{lightGreen}  
12 & Filter & a passive band-pass filter & \cellcolor{lightRed}26 & \cellcolor{lightRed}Adder & \cellcolor{lightRed}Op-map adder \\
\rowcolor{lightGreen} 
13 & Filter & a passive band-stop filter & \cellcolor{lightRed}27 & \cellcolor{lightRed}Subtractor & \cellcolor{lightRed}Op-map substrator \\
\rowcolor{lightOrange} 
14 & Amplifier & Common-source amp. with diode-connected load & \cellcolor{lightRed}28 & \cellcolor{lightRed}Schmitt trigger & \cellcolor{lightRed}Non-inverting Schimiit trigger \\
\bottomrule
\end{tabular}
\end{adjustbox}
\end{table*}

\textbf{Hyper-parameters.}
For topology generation, we set the temperature to 0.5 and top\_p to 1.0 to encourage output diversity, while for parameter extraction, the temperature is fixed at 0.0.
For circuit device sizing, we use Bayesian optimization with the Optuna package~\cite{akiba2019optuna} (TPE sampler, 1000 trials), including 25\% random initialization (250 trials) and multivariate sampling to account for parameter dependencies. 

The feedback flow allows up to three attempts, balancing token cost and success rate since most errors are corrected within this range.
For multimodal LLMs, we use the native version of each model when available (e.g., GPT‑4.1); otherwise, we employ alternative variants (GPT‑4o for DeepSeek‑V3, Qwen‑VL for Qwen‑Coder). 
All experiments are conducted with the BPTM 45nm PDK, consistent with AnalogGenie~\cite{gao2025analoggenielite}.

\begin{figure}[!t]
    \centering
        \centering
        \includegraphics[width=\columnwidth]{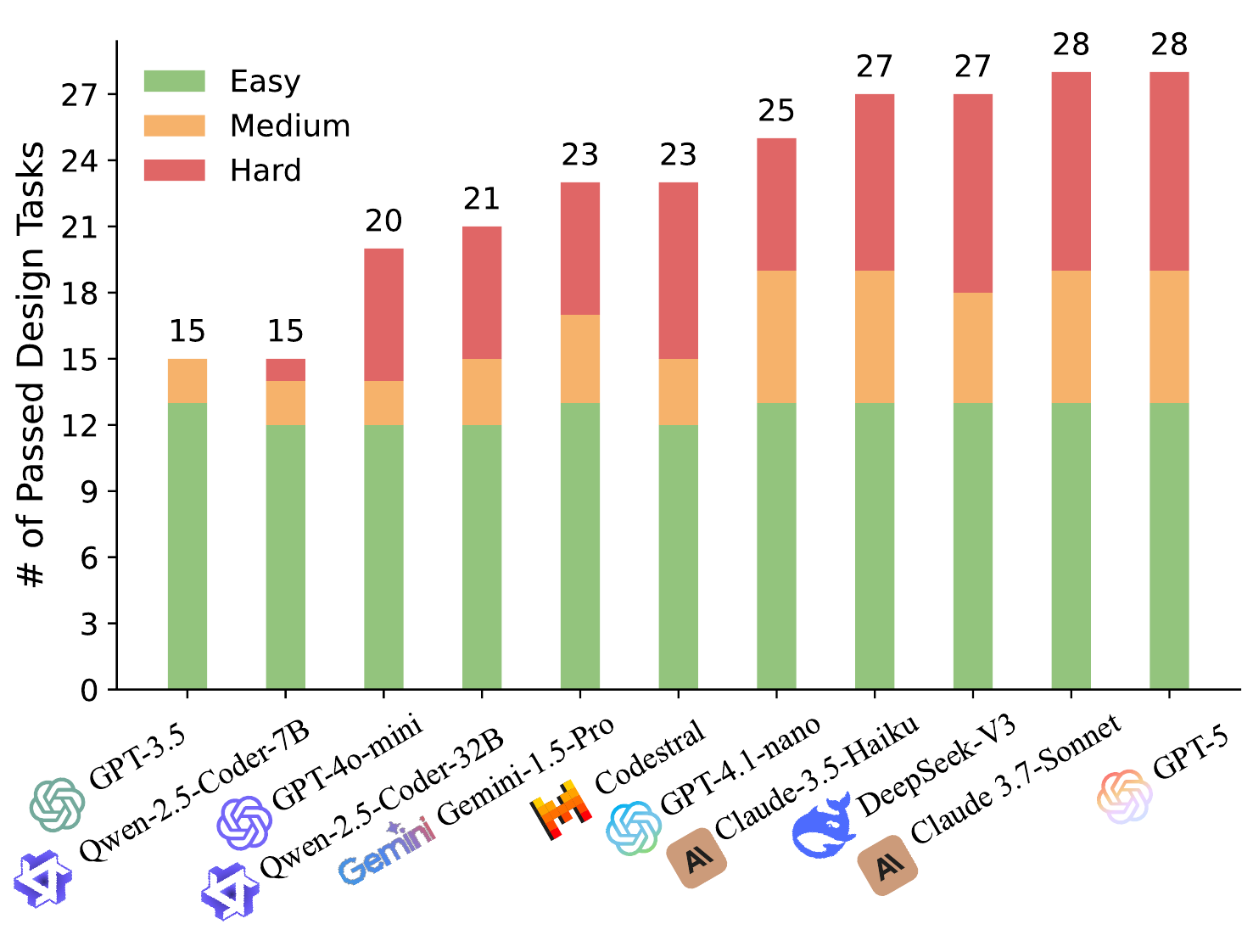} 
        \caption{\textbf{Leaderboard of LLM analog circuit design.} 
        }
        \label{leaderboard}
    \vspace{-10pt}
\end{figure}

\subsection{Main Results}

For the topology generation task, we evaluate the functional correctness of the generated netlists. Each task is attempted 30 times per model and counted as solved if at least one valid netlist is produced.
Fig.~\ref{leaderboard} summarizes the number of solved tasks as a leaderboard.
In just over two years, the number of circuits successfully designed by LLMs doubled from 14 with GPT‑3.5 to 28 with GPT‑5, highlighting the rapid advancement of LLMs in circuit design.
Table~\ref{main_result} provides a quantitative comparison of multiple LLMs evaluated in the same AnalogCoder‑Pro framework under identical settings.
The results indicate that the best proprietary models outperform open‑source alternatives, with GPT‑5 achieving the best performance (28/28 tasks, 98.6\% Pass@5). 
Moreover, there is a clear positive correlation between general‑purpose LLM capability and analog circuit generation performance, with more advanced models consistently achieving higher success rates across all circuit types.

\begin{table*}[!t]
\centering
\caption{{\textbf{Main results.} All LLMs are evaluated using the complete AnalogCoder-Pro framework}}
\begin{adjustbox}{max width=\textwidth}
\begin{tabular}{r|l|rr|rr|rr|rr|rr|rr|rr}
\toprule
\textbf{Model} & \multirow{2}{*}{\textbf{Circuit Type}} & \multicolumn{2}{c|}{\textbf{Qwen-2.5-Coder-32B}} & \multicolumn{2}{c|}{\textbf{GPT-4.1-nano}} & \multicolumn{2}{c|}{\textbf{Gemini-2.5-Flash}} & \multicolumn{2}{c|}{\textbf{DeepSeek-V3}} & \multicolumn{2}{c|}{\textbf{GPT-4.1}} & \multicolumn{2}{c|}{\textbf{Claude-3.7-Sonnet}} & \multicolumn{2}{c}{\textbf{GPT-5}} \\
Task ID & & Pass@1 & Pass@5 & Pass@1 & Pass@5 & Pass@1 & Pass@5 & Pass@1 & Pass@5 & Pass@1 & Pass@5 & Pass@1 & Pass@5 & Pass@1 & Pass@5 \\
\midrule
1  & Amplifier & 80.0 & 100.0 & 73.3 & 100.0 & 100.0 & 100.0 & 100.0 & 100.0 & 100.0 & 100.0 & 100.0 & 100.0 & 100.0 & 100.0 \\
2  & Amplifier & 80.0 & 100.0 & 6.7 & 31.0 & 100.0 & 100.0 & 100.0 & 100.0 & 100.0 & 100.0 & 100.0 & 100.0 & 100.0 & 100.0 \\
3  & Amplifier & 70.0 & 99.9 & 33.3 & 89.1 & 100.0 & 100.0 & 100.0 & 100.0 & 100.0 & 100.0 & 100.0 & 100.0 & 100.0 & 100.0 \\
4  & Amplifier & 66.7 & 99.8 & 90.0 & 100.0 & 80.0 & 100.0 & 96.7 & 100.0 & 100.0 & 100.0 & 100.0 & 100.0 & 100.0 & 100.0 \\
5  & Amplifier & 20.0 & 70.2 & 26.7 & 81.5 & 93.3 & 100.0 & 100.0 & 100.0 & 100.0 & 100.0 & 100.0 & 100.0 & 100.0 & 100.0 \\
6  & Inverter & 100.0 & 100.0 & 86.7 & 100.0 & 100.0 & 100.0 & 100.0 & 100.0 & 100.0 & 100.0 & 100.0 & 100.0 & 100.0 & 100.0 \\
7  & Inverter & 100.0 & 100.0 & 86.7 & 100.0 & 100.0 & 100.0 & 100.0 & 100.0 & 100.0 & 100.0 & 100.0 & 100.0 & 100.0 & 100.0 \\
8  & Current Mirror & 73.3 & 100.0 & 30.0 & 85.7 & 76.7 & 100.0 & 100.0 & 100.0 & 100.0 & 100.0 & 100.0 & 100.0 & 100.0 & 100.0 \\
9  & Comparator & 53.3 & 98.6 & 73.3 & 100.0 & 3.3 & 16.7 & 100.0 & 100.0 & 100.0 & 100.0 & 100.0 & 100.0 & 53.3 & 98.6 \\
10 & Filter & 80.0 & 100.0 & 93.3 & 100.0 & 96.7 & 100.0 & 100.0 & 100.0 & 96.7 & 100.0 & 100.0 & 100.0 & 100.0 & 100.0 \\
11 & Filter & 13.3 & 53.8 & 86.7 & 100.0 & 100.0 & 100.0 & 100.0 & 100.0 & 90.0 & 100.0 & 100.0 & 100.0 & 100.0 & 100.0 \\
12 & Filter & 16.7 & 62.7 & 13.3 & 53.8 & 100.0 & 100.0 & 86.7 & 100.0 & 80.0 & 100.0 & 100.0 & 100.0 & 96.7 & 100.0 \\
13 & Filter & 0.0 & 0.0 & 6.7 & 31.0 & 100.0 & 100.0 & 36.7 & 91.8 & 36.7 & 91.8 & 66.7 & 100.0 & 100.0 & 100.0 \\
14 & Amplifier & 43.3 & 95.7 & 36.7 & 91.8 & 26.7 & 81.5 & 100.0 & 100.0 & 100.0 & 100.0 & 100.0 & 100.0 & 100.0 & 100.0 \\
15 & Amplifier & 30.0 & 85.7 & 66.7 & 99.8 & 100.0 & 100.0 & 100.0 & 100.0 & 100.0 & 100.0 & 100.0 & 100.0 & 100.0 & 100.0 \\
16 & Current Mirror & 0.0 & 0.0 & 16.7 & 62.7 & 100.0 & 100.0 & 100.0 & 100.0 & 76.7 & 100.0 & 100.0 & 100.0 & 96.7 & 100.0 \\
17 & Opamp & 0.0 & 0.0 & 3.3 & 16.7 & 30.0 & 85.7 & 60.0 & 99.4 & 66.7 & 99.8 & 6.7 & 33.3 & 66.7 & 99.8 \\
18 & Opamp & 0.0 & 0.0 & 16.7 & 62.7 & 96.7 & 100.0 & 96.7 & 100.0 & 93.3 & 100.0 & 100.0 & 100.0 & 100.0 & 100.0 \\
19 & Mixer & 20.0 & 70.2 & 6.7 & 31.0 & 3.3 & 16.7 & 66.7 & 99.8 & 50.0 & 97.9 & 100.0 & 100.0 & 96.7 & 100.0 \\
20 & Opamp & 6.7 & 31.0 & 0.0 & 0.0 & 100.0 & 100.0 & 76.7 & 100.0 & 90.0 & 100.0 & 100.0 & 100.0 & 100.0 & 100.0 \\
21 & Opamp & 0.0 & 0.0 & 0.0 & 0.0 & 90.0 & 100.0 & 6.7 & 31.0 & 46.7 & 96.9 & 86.7 & 100.0 & 56.7 & 99.1 \\
22 & Oscillator & 16.7 & 62.7 & 16.7 & 62.7 & 66.7 & 99.8 & 10.0 & 43.4 & 46.7 & 96.9 & 46.7 & 98.1 & 16.7 & 62.7 \\
23 & Oscillator & 0.0 & 0.0 & 0.0 & 0.0 & 76.7 & 100.0 & 30.0 & 85.7 & 73.3 & 100.0 & 100.0 & 100.0 & 100.0 & 100.0 \\
24 & Integrator & 13.3 & 53.8 & 80.0 & 100.0 & 100.0 & 100.0 & 96.7 & 100.0 & 96.7 & 100.0 & 100.0 & 100.0 & 100.0 & 100.0 \\
25 & Differentiator & 30.0 & 85.7 & 40.0 & 94.0 & 100.0 & 100.0 & 100.0 & 100.0 & 100.0 & 100.0 & 100.0 & 100.0 & 100.0 & 100.0 \\
26 & Adder & 60.0 & 99.4 & 63.3 & 99.7 & 100.0 & 100.0 & 100.0 & 100.0 & 30.0 & 85.7 & 100.0 & 100.0 & 80.0 & 100.0 \\
27 & Subtractor & 0.0 & 0.0 & 26.7 & 81.5 & 93.3 & 100.0 & 93.3 & 100.0 & 73.3 & 100.0 & 100.0 & 100.0 & 96.7 & 100.0 \\
28 & Schmitt trigger & 20.0 & 70.2 & 20.0 & 70.2 & 93.3 & 100.0 & 40.0 & 94.0 & 33.3 & 89.1 & 66.7 & 100.0 & 90.0 & 100.0 \\
\midrule
Avg & & 35.5 & 62.1 & 39.3 & 69.5 & 83.1 & 92.9 & 82.0 & 94.5 & 81.4 & 98.5 & \textbf{91.9} & 97.5 & 91.1 & \textbf{98.6} \\
\# Solved & & 21 & 21 & 25 & 25 & \textbf{28} & \textbf{28} & \textbf{28} & \textbf{28} & \textbf{28} & \textbf{28} & \textbf{28} & \textbf{28} & \textbf{28} & \textbf{28} \\
\bottomrule
\end{tabular}
\end{adjustbox}
\label{main_result}
\end{table*}

\begin{table}[!ht]
\centering
\caption{{\textbf{Ablation Study Results.} }}
\begin{adjustbox}{width=\linewidth}
\label{ablation}
\begin{tabular}{l|l|r|r|r}
\toprule
\textbf{Backbone} & \textbf{Ablation} & \textbf{Pass@1} & \textbf{Pass@5} & \textbf{\# Solved} \\
\midrule
\multirow{6}{*}{\textbf{GPT-4.1-nano}} 
& SPICE netlist          &  14.4 & 33.2  & 20 \\
& w/o in-context learning     & 25.8  & 47.3  & 22 \\
& w/o chain-of-thought      &  25.0 & 50.8  & 22 \\
& w/o feedback flow     & 18.8  & 45.0  & 19 \\
& w/o library tool      & 24.0  & 45.1  & 20 \\
& full framework     &  \textbf{39.3} & \textbf{69.5}  & \textbf{25} \\
\midrule
\multirow{6}{*}{\textbf{DeepSeek-V3}} 
& SPICE netlist          & 77.6  & 89.8  & 27 \\
& w/o in-context learning           & 64.2  & 83.0  & 27 \\
& w/o chain-of-thought           & 75.2  & 89.4  & 26 \\
& w/o feedback flow          & 69.8  & 80.3  & 27 \\
& w/o library tool          & 75.0  & 84.8  & 24 \\
& full framework    & \textbf{82.0}  & \textbf{94.5}  & \textbf{28} \\
\bottomrule
\end{tabular}
\end{adjustbox}
\end{table}

\begin{table}[!ht]
\centering
\caption{\textbf{Comparison of Multimodal Feedback-Enhanced Flow.} 
Each model is evaluated with and without Multimodal LLM assistance (with $\bullet$ denoting assistance) in average Pass@1. 
The \textbf{Avg.} reports the mean score across categories. 
Circuit type abbreviations follow Table~\ref{tab:circuit_simulation}.}
\label{tab:mllm_debug}
\begin{adjustbox}{width=\linewidth}
\begin{tabular}{l|c|rrrrrrr|r}
\toprule
\textbf{Model} & \textbf{Multimodal} & \textbf{Osc} & \textbf{Int} & \textbf{Diff} & \textbf{Sch} & \textbf{Comp} & \textbf{Mix} & \textbf{Filt} & \textbf{Avg.} \\
\midrule
Qwen2.5 & & 1.7 & 6.7 & 40.0 & 20.0 & 20.0 & 16.7 & 16.7 & 17.4 \\
-Coder-32B & $\bullet$ & 8.3 & 13.3 & 30.0 & 20.0 & 53.3 & 20.0 & 27.5 & \textbf{24.6} \\
\midrule
\multirow{2}{*}{GPT-4.1} & & 43.3 & 93.3 & 96.7 & 36.7 & 93.3 & 43.3 & 75.0 & 68.8 \\
& $\bullet$ & 60.0 & 96.7 & 100.0 & 33.3 & 100.0 & 50.0 & 75.8 & \textbf{73.7} \\

\midrule
\multirow{2}{*}{DeepSeek-V3} & & 8.3 & 100.0 & 96.7 & 50.0 & 100.0 & 63.3 & 77.5 & 70.8 \\
& $\bullet$ & 20.0 & 96.7 & 100.0 & 40.0 & 100.0 & 66.7 & 80.8 & \textbf{72.0} \\
\midrule
Claude-3.7 & & 70.0 & 100.0 & 100.0 & 26.7 & 100.0 & 100.0 & 91.7 & 84.1 \\
-Sonnet & $\bullet$ & 73.3 & 100.0 & 100.0 & 66.7 & 100.0 & 100.0 & 91.7 & \textbf{90.2} \\
\bottomrule
\end{tabular}
\end{adjustbox}
\end{table}

\subsection{Unified Generation and Optimization}

For each circuit type, we compared two prompt strategies: one including an explicit design objective (e.g., ``Design a multi‑stage op‑amp topology that optimizes FoM (GBW*CL/Power)'') and one omitting it (e.g., ``Design a multi‑stage op‑amp''), to evaluate the effectiveness of unified design.
Optimization is performed with 1000 samples per trial, and the best value from each run is recorded. The unified generation and optimization process is repeated 30 times, and we report the best value among all 30 trials. 
As shown in Table~\ref{tab:comparison}, each base model is reported with two rows (without vs. with target-guided prompts), and the target-guided setting consistently improves objective-specific results, demonstrating that the LLM can effectively incorporate optimization goals during topology generation.

\begin{figure}[!ht]
\centering
\includegraphics[width=\columnwidth]{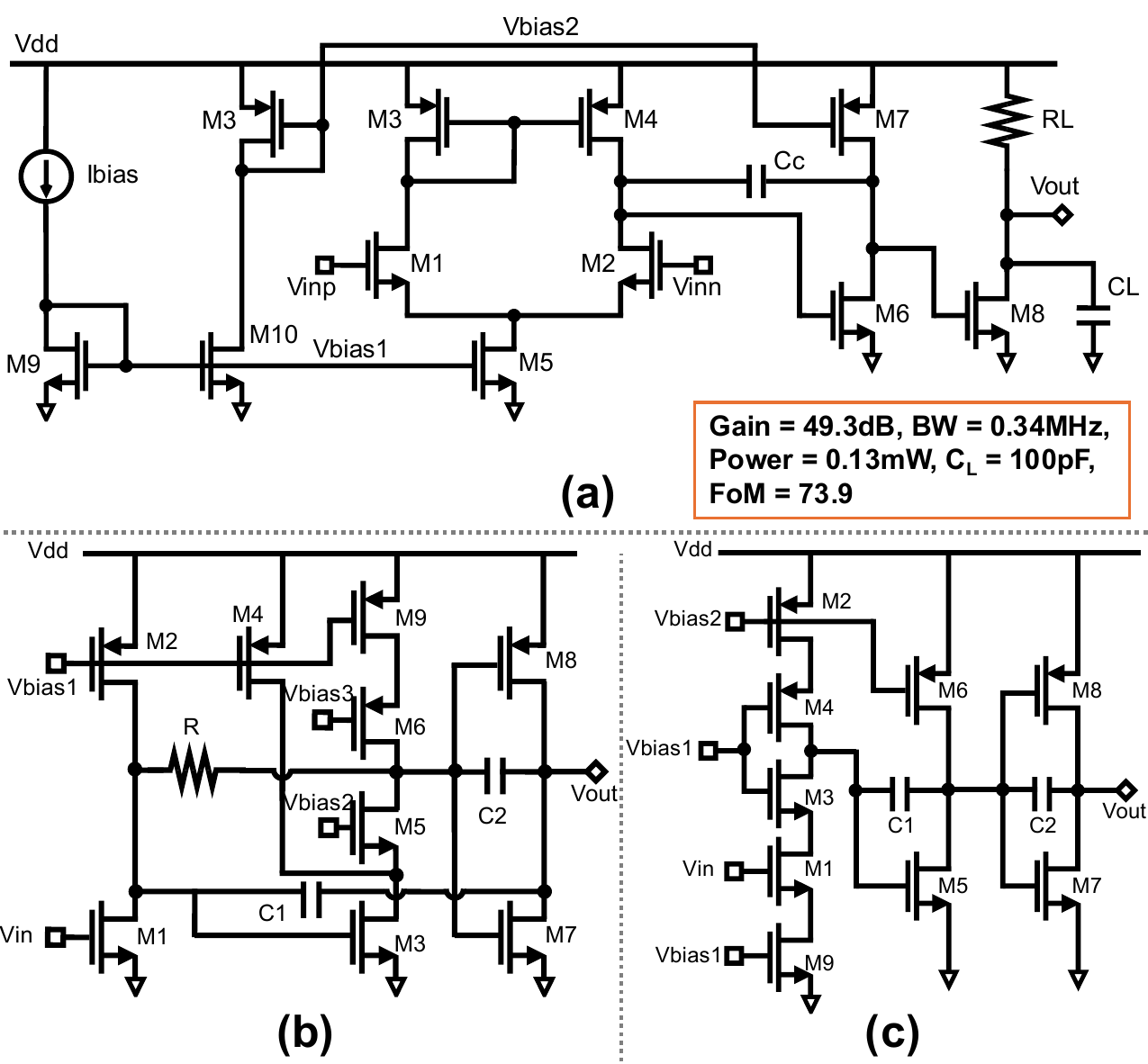}
\caption{
\textbf{Topologies generated by AnalogCoder-Pro.} 
(a) Op-amp with the highest FoM, also evaluated in Fig.~\ref{fig:input_bias_determination}; 
(b), (c) Amplifiers generated with target-guided design prompts.
}
\label{fig:generated_circuit}
\end{figure}

\begin{figure*}[!th]
\centering
\includegraphics[width=\textwidth]{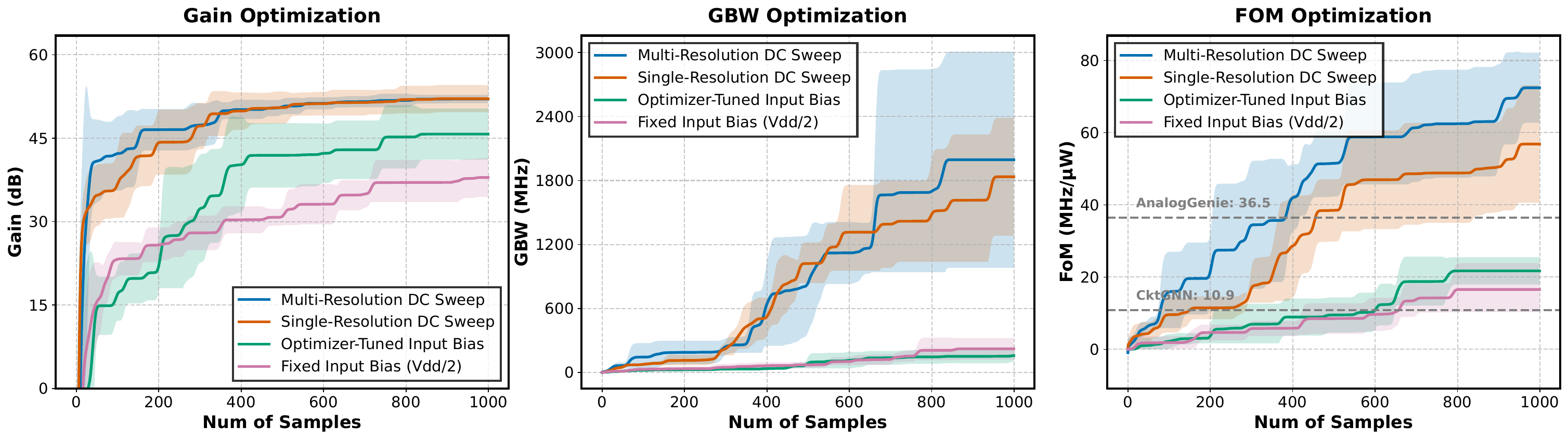}
\caption{
\textbf{Comparison of Input Bias Voltage Selection Methods.}
Each method is evaluated over 5 independent optimization runs. 
The proposed multi-resolution bias selection approach achieves faster and more accurate convergence to the optimal operating points, resulting in improved circuit performance metrics. 
The circuit topology for evaluation is shown in Fig.~\ref{fig:generated_circuit}(a).
}
\label{fig:input_bias_determination}
\vspace{-10pt}
\end{figure*}

Beyond the analysis of unified prompt strategies, we further evaluate the performance of AnalogCoder-Pro in terms of FoM. 
According to Table~\ref{tab:comparison}, AnalogCoder, aided by multi-resolution input bias search, generates circuits that achieve competitive baseline performance even with only the initial LLM-assigned sizes and no further optimization. 
In contrast, AnalogCoder-Pro combines LLM-based parameter extraction with Bayesian optimization, yielding substantial gains over AnalogGenie, including up to $2\times$ better op-amp FoM. 
To directly assess the role of multi-resolution input bias search, we compared it with several alternatives, including single-resolution search, fixed bias, and treating input bias as an optimizable parameter, using the best FoM netlist generated by Claude-3.7 (Fig.~\ref{fig:generated_circuit}(a)).
For each method, the optimizer was run using five random seeds, constraining the input bias range to $0.25V_\mathrm{dd}$--$0.75V_\mathrm{dd}$ to avoid supply rail nonlinearities; variance across runs is shown as error bars. 
The single-resolution baseline employed 2000 uniform steps, while the multi-resolution strategy used 20, 200, and 2000 steps at coarse, medium, and fine stages. 
As shown in Fig.~\ref{fig:input_bias_determination}, the multi-resolution search consistently outperforms all baselines across every evaluation metric. 
Additional amplifier topologies generated by AnalogCoder-Pro are provided in Fig.~\ref{fig:generated_circuit}(b) and (c).
\looseness=-1

\begin{table*}[!th]
\centering{
\caption{
\textbf{Performance Comparison for Unified Circuit Generation and Optimization.}
For each instruction in Fig.~\ref{unified_design_task}, 30 netlist generation attempts are performed. Each correct netlist is optimized with 1000 samples, and the best performance among the 30 trials is reported. Maximum values are in \textbf{bold}. AnalogGenie and CktGNN results are from the original papers.
}
\label{tab:comparison}
\begin{adjustbox}{width=\textwidth}
\begin{tabular}{c|l|c|rrr|rrr|rrr|rrr|r}
\toprule
\textbf{Method} & \multirow{2}{*}{\textbf{Base Model}} & \textbf{Target-} & \multicolumn{3}{c|}{\textbf{Single-Stage Amp}} & \multicolumn{3}{c|}{\textbf{Single-Stage Opamp}} & \multicolumn{3}{c|}{\textbf{Multi-Stage Amp}} & \multicolumn{3}{c|}{\textbf{Multi-Stage Opamp}} & \textbf{Avg.} \\
\textbf{\# of Simulations}& & \textbf{guided}\textsuperscript{2} & \textbf{Gain} & \textbf{GBW} & \textbf{FoM}\textsuperscript{1} & \textbf{Gain} & \textbf{GBW} & \textbf{FoM} & \textbf{Gain} & \textbf{GBW} & \textbf{FoM} & \textbf{Gain} & \textbf{GBW} & \textbf{FoM} & \textbf{Rank} \\
\midrule
\textbf{CktGNN}~\cite{dong2023cktgnn} & GNN & $\bullet$ & \text{-} & \text{-} & \text{-} & \text{-} & \text{-} & \text{-} & \text{-} & \text{-} & \text{-} & \text{-} & \text{-} & 10.9 & - \\
\midrule
\textbf{AnalogGenie($>$1k)}~\cite{gao2025analoggenie} & GPT-2 & $\bullet$ & \text{-} & \text{-} & \text{-} & \text{-} & \text{-} & \text{-} & \text{-} & \text{-} & \text{-} & \text{-} & \text{-} & 36.5 & - \\
\midrule
\multirow{6}{*}{\textbf{AnalogCoder(1)}~\cite{lai2025analogcoder}} 
& \multirow{2}{*}{GPT-4o} &  & 13.9 & 93.2 & 1.7 & 12.8 & 113.1 & 0.6 & 40.6 & 2076.2 & 5.7 & 18.0 & 154.6 & 0.5 & 10.5 \\
& & $\bullet$ & 16.2 & 191.7 & 1.1 & 13.4 & 85.1 & 0.8 & 43.8 & 4195.4 & 8.8 & 35.5 & 880.8 & 2.4 & 9.1 \\
\cmidrule{2-16}
& \multirow{2}{*}{DeepSeek-V3} &  & 0.7 & 23.6 & 1.7 & 13.5 & 147.7 & 0.9 & 37.5 & 5330.6 & 8.5 & 27.8 & 1050.9 & 4.1 & 9.4 \\
& & $\bullet$ & 26.2 & 198.5 & 1.0 & 29.7 & 153.5 & 2.6 & 37.5 & 9808.7 & 13.1 & 7.8 & 443.1 & 4.2 & 7.6 \\
\cmidrule{2-16}
& \multirow{2}{*}{Claude-3.7} &  & 16.0 & 49.4 & 1.9 & 14.5 & 130.0 & 1.1 & 35.8 & 314.8 & 1.2 & 67.9 & 2384.8 & 12.1 & 8.2 \\
& & $\bullet$ & 30.5 & 157.4 & 1.0 & 23.1 & 13.7 & 1.1 & 0.0 & 4081.3 & 165.9 & 40.0 & 899.0 & 0.5 & 7.9 \\
\midrule
\multirow{6}{*}{\textbf{AnalogCoder-Pro(1k)}} 
& \multirow{2}{*}{GPT-4o} &  & 18.6 & 416.8 & 2.0 & 20.6 & 299.1 & 2.9 & 56.3 & 23111.9 & 133.2 & 33.3 & 1406.1 & 9.5 & 4.9 \\
& & $\bullet$ & 18.9 & \textbf{443.3} & 2.4 & 20.4 & 260.3 & 3.6 & 55.7 & 35873.3 & 124.0 & 50.2 & 1122.2 & 37.4 & 4.5 \\
\cmidrule{2-16}
& \multirow{2}{*}{DeepSeek-V3} &  & 16.1 & 50.4 & 1.9 & 20.5 & 279.9 & 3.0 & 55.9 & 22579.2 & 127.3 & 36.4 & 2346.0 & 14.1 & 5.7 \\
& & $\bullet$ & 27.5 & 418.8 & 2.5 & 36.2 & \textbf{363.3} & \textbf{14.5} & \textbf{56.4} & \textbf{41316.7} & 185.7 & 36.6 & 2243.9 & 36.2 & \textbf{2.5} \\
\cmidrule{2-16}
& \multirow{2}{*}{Claude-3.7} &  & 16.1 & 100.6 & 1.9 & 27.8 & 281.3 & 8.5 & 53.2 & 5298.2 & 135.1 & 67.9 & \textbf{6279.4} & 43.4 & 4.3 \\
& & $\bullet$ & \textbf{45.3} & 413.3 & \textbf{3.2} & \textbf{39.2} & 266.6 & 7.4 & 46.1 & 17366.0 & \textbf{413.2} & \textbf{74.9} & 3193.0 & \textbf{73.9} & 2.6 \\
\bottomrule
\multicolumn{16}{l}{\textsuperscript{1}\ {Gain (dB) = $20 \times \log_{10}(V_\text{out}/V_\text{in})$, \  GBW (MHz) = Gain (linear) $\times$ Bandwidth, \  $\mathrm{FoM}$ (MHz$\cdot$pF/$\mu$W) $= \mathrm{GBW} \times C_L/\mathrm{Power}$, where $C_L=100$\,pF.
}} \\
\multicolumn{16}{l}{\textsuperscript{2}\  {Target-guided design indicates that specific design objectives (e.g., maximizing the Figure of Merit, FoM) were set for the LLM, as shown in Fig.~\ref{unified_design_task}.
}
}
\end{tabular}
\end{adjustbox}
}
\vspace{-10pt}
\end{table*}

\subsection{Ablation study}

Our ablation study investigates the contribution of different components in the AnalogCoder-Pro framework. Specifically, we examine the effects of (i) having the LLM directly generate SPICE netlists instead of Python code, (ii) removing in-context learning from the design prompts, (iii) removing chain-of-thought guidance from the design prompts, (iv) excluding the feedback‑enhanced workflow, and (v) omitting the circuit tool library, in comparison with (vi) the full framework.
The results in Table~\ref{ablation} show that removing any component reduces performance. 
To further evaluate the role of MLLM, we compare the results with and without multimodal waveform analysis in the feedback flow. 
As shown in Table~\ref{tab:mllm_debug}, using MLLM consistently improves circuit topology generation success rate.

\section{Conclusion}
\label{conclusion}

In this work, we introduce AnalogCoder-Pro, a unified, end-to-end multimodal LLM framework for automated analog circuit design. By combining multimodal diagnosis-and-repair mechanisms with a reusable circuit tool library and integrated performance-driven topology generation, AnalogCoder-Pro achieves state-of-the-art performance and substantially improves success rates across diverse circuit design tasks.


\bibliographystyle{unsrt}
\bibliography{refs}

\end{document}